\documentclass[10pt,twocolumn,letterpaper]{article}

\usepackage[pagenumbers]{cvpr} 

\usepackage{times}
\usepackage{epsfig}
\usepackage{graphicx}
\usepackage{amsmath}
\usepackage{amssymb}
\usepackage{booktabs}
\usepackage{multirow}
\usepackage{textcomp}
\usepackage{xcolor}
\usepackage{url}

\definecolor{cvprblue}{rgb}{0.21,0.49,0.74}
\usepackage[pagebackref,breaklinks,colorlinks,allcolors=cvprblue]{hyperref}

\def\paperID{*****}
\def\confName{CVPR}
\def\confYear{2026}

\title{Zero-Shot Cross-Material Ptychographic Phase Reconstruction Using Deep Learning}

\author{
Wen-Chun Lin$^{1}$ \qquad
Kuan-Peng Chen$^{2}$ \qquad
Nan-Yow Chen$^{2}$ \\
Yu-Chee Tseng$^{1}$ \qquad
Jen-Jee Chen$^{1}$\\
$^{1}$Institute of Computational Intelligence, College of Artificial Intelligence,\\
National Yang Ming Chiao Tung University, Taiwan\\
$^{2}$National Center for High-performance Computing,\\
National Institutes of Applied Research, Taiwan\\
}

\begin{document}
\maketitle

\begin{abstract}
Ptychographic phase reconstruction is commonly formulated as an iterative inverse problem, requiring repeated object–probe updates and resulting in substantial computational cost for large-scale 4D-STEM data. We present a \textbf{direct local-to-global learning framework} that reconstructs full-field phase maps from diffraction measurements without iterative refinement during inference. The proposed network predicts local wrapped-phase patches from individual diffraction patterns using a sine–cosine representation, and the predictions are assembled into a full-field reconstruction using calibrated scan positions and Gaussian-weighted stitching. To evaluate generalization beyond the training domain, the model is trained on one material and directly applied to another in a \textbf{zero-shot setting without target-domain fine-tuning.} Experiments on AuPd and MoS$_2$ demonstrate consistent \textbf{cross-material transfer in both directions}, with the proposed method achieving the best full-field MSE, PSNR, and MS-SSIM among the evaluated learning-based methods. Compared with the iterative ePIE approach, the proposed direct local-to-global pipeline reduces end-to-end reconstruction time by approximately \textbf{10$\times$}, demonstrating its potential for efficient and transferable ptychographic reconstruction.
\end{abstract}

\section{Introduction}

Ptychography is a coherent diffractive imaging technique in which a localized probe is scanned across a specimen and diffraction patterns are recorded at overlapping positions. Although the detector measures only intensity, the redundancy introduced by overlapping measurements provides strong constraints for recovering the complex object information \cite{rodenburg2004,maiden2009}. In electron ptychography and four-dimensional scanning transmission electron microscopy (4D-STEM), a two-dimensional diffraction pattern is acquired at each two-dimensional scan position, forming four-dimensional data from which the object phase can be reconstructed \cite{yu2024,ophus2019}. The recovered phase provides structural and electrostatic-potential information that is valuable for materials characterization but is not directly available from intensity-only images.

Ptychographic reconstruction is commonly formulated as an \textit{iterative} inverse problem. Classical methods, including the ptychographical iterative engine (PIE), the extended ptychographical iterative engine (ePIE), and difference-map-based algorithms, repeatedly update the object and probe estimates to obtain a solution consistent with the measured diffraction intensities \cite{rodenburg2004,maiden2009,thibault2009}. Although effective, these methods require \textit{repeated} forward and backward operations and can therefore be computationally expensive for large-scale 4D-STEM data cubes, particularly when many scan positions or extensive parameter tuning are involved \cite{yu2024,ophus2019}. \textit{Direct} approaches, such as Wigner distribution deconvolution, have been developed to address the latency issue and support high-throughput processing \cite{yang2017,bangun2023}.

Deep learning offers an alternative by learning an approximate inverse mapping from diffraction measurements to object phase. Neural, physics-informed, hybrid, transformer-based, and pre-trained approaches have demonstrated the potential to accelerate or complement iterative reconstruction \cite{cherukara2020,hoidn2023,yamada2024,nakahata2024,yue2025,pan2023,chang2023}. However, learned reconstruction models may rely on material- or acquisition-specific characteristics in the training data. 
These methods struggle with \textit{distribution shift}, especially when a model trained on one material is applied to another \cite{ongie2020}. Our objective is to enable \textit{zero-shot transfer} across distinct material domains for learning-based electron ptychography.

In this work, we formulate ptychographic phase retrieval as a \textbf{phase-only local-to-global} feed-forward reconstruction problem with an emphasis on \textbf{zero-shot cross-material generalization}. The network predicts local wrapped-phase patches from individual diffraction patterns, and the local predictions are directly assembled into a full-field phase map using calibrated scan coordinates. To assess cross-material transferability, we train our model on one material and evaluate it on another without target-domain training or fine-tuning.

\begin{figure*}[t]
    \centering
    \IfFileExists{figures/fig1_overall_workflow.png}{%
        \includegraphics[width=0.98\textwidth]{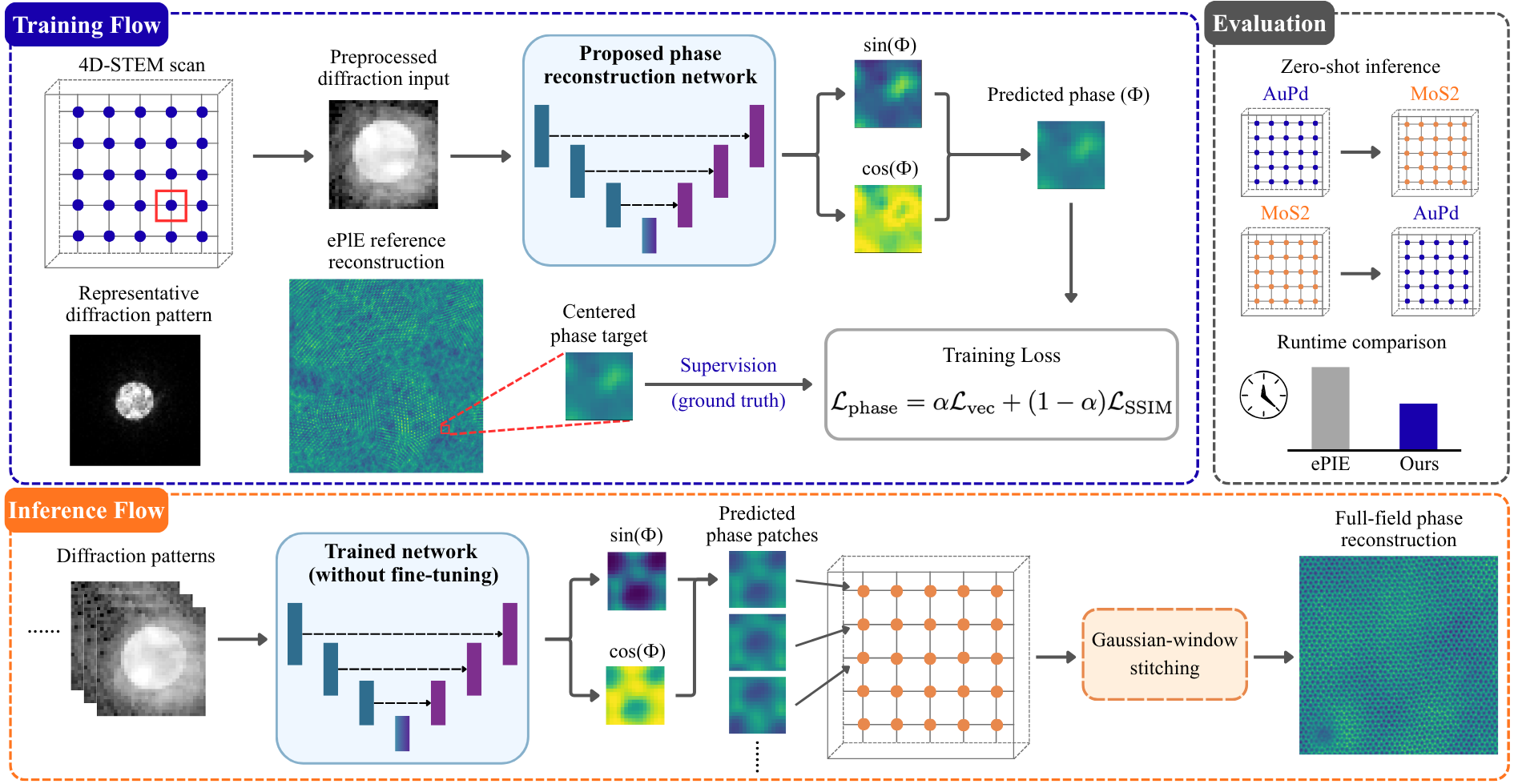}%
    }{%
        \fbox{\parbox[c][1.65in][c]{0.95\textwidth}{\centering Placeholder for Fig.~1. Put your workflow image at \texttt{figures/fig1\_overall\_workflow.png}.}}%
    }
    \caption{Overview of the proposed zero-shot cross-material ptychographic reconstruction framework, from local sine--cosine phase prediction to Gaussian-window full-field assembly.}
    \label{fig:workflow}
\end{figure*}

The main contributions of this work are threefold. First, we investigate \textbf{zero-shot cross-material ptychographic reconstruction} by training on one material and evaluating directly on another without target-domain fine-tuning. This setting examines whether the learned diffraction-to-phase mapping captures \textbf{transferable structural information} rather than only material-specific characteristics. Second, we investigate the \textbf{local supervision scale} and show that centered $16\times16$ phase targets provide a compact and empirically effective prediction unit across the evaluated transfer settings. Under the calibrated object-space sampling, each target spans approximately $3.10$~\AA, giving the local prediction unit a physically interpretable spatial extent. Third, we develop a phase-only local-to-global reconstruction pipeline that combines periodic \textbf{sine--cosine prediction} with scan-position-aware \textbf{Gaussian-window circular stitching}, enabling full-field phase reconstruction without iterative object--probe updates during inference.

\section{Related Work}

Ptychographic reconstruction has traditionally relied on model-based algorithms. PIE and ePIE iteratively update object and probe estimates, while difference-map and probe-retrieval variants jointly recover the specimen transmission function and illumination profile \cite{rodenburg2004,maiden2009,thibault2009}. In electron ptychography, 4D-STEM provides overlapping diffraction measurements for quantitative phase reconstruction, and direct schemes such as single-sideband reconstruction and Wigner distribution deconvolution (WDD) reduce latency relative to iterative solvers \cite{ophus2019,yu2024,yang2017,bangun2023}. However, this computational efficiency is generally accompanied by a stronger dependence on acquisition and calibration information, such as scan sampling, probe convergence angle, reciprocal-space calibration, and the relative orientation between scan and detector coordinates. WDD may additionally depend on an appropriate probe transfer model and regularization for stable deconvolution. Consequently, although these direct methods avoid costly iterative optimization, reliable reconstruction still requires accurate experimental geometry and calibration, which may limit their applicability when acquisition metadata are incomplete or experimental conditions vary. These methods nevertheless establish important physical and computational baselines for ptychographic reconstruction.

Learning-based methods instead approximate the inverse mapping from diffraction measurements to object information. PtychoNN demonstrated fast convolutional reconstruction from far-field diffraction data \cite{cherukara2020}, while Friedrich et al.~\cite{friedrich2023} recovered local phase-object patches from neighboring 4D-STEM diffraction patterns and combined them into a full phase object. Subsequent work has incorporated diffraction physics and overlap constraints through physics-informed learning \cite{hoidn2023}, combined neural predictions with iterative refinement \cite{yamada2024}, and used transformer architectures to model broader dependencies among diffraction measurements \cite{nakahata2024,yue2025}. Pre-training and fine-tuning have also been investigated to improve reconstruction quality and transferability \cite{pan2023}, while related studies have extended learned inversion to high-resolution electron-scattering reconstruction and real-time edge deployment \cite{chang2023,babu2023}. These approaches differ in supervision, physical constraints, and local or global prediction scope, but all seek to reduce the cost of iterative reconstruction.

Prior work has examined generalization through out-of-distribution objects, unseen probes or scan patterns, and target-domain fine-tuning \cite{hoidn2023,nakahata2024,pan2023}; however, explicit zero-shot transfer between distinct material domains in electron ptychography has received comparatively limited attention. Closest to our local-to-global design, the method in~\cite{friedrich2023} reconstructs local phase-object patches and subsequently combines them into a global reconstruction. Our study differs by focusing on phase-only cross-material transfer without target-domain fine-tuning, using periodic sine--cosine prediction and calibrated Gaussian-weighted circular stitching to test whether local diffraction-to-phase mappings remain transferable across materials.

\section{Method}

\subsection{Problem Formulation}

In a 4D-STEM ptychographic experiment, a focused electron probe is raster-scanned across a specimen on a two-dimensional grid. At each scan position, the detector records a two-dimensional diffraction-intensity pattern. Neighboring probe positions illuminate overlapping regions of the specimen, providing redundant measurements. The complete set of diffraction patterns is then used to reconstruct the specimen’s spatial phase map.

We formulate phase reconstruction as a \textbf{local-to-global learning} problem. Rather than estimating the complete phase map from all diffraction measurements simultaneously, the objective is to learn a shared mapping from an individual diffraction pattern to a local phase patch associated with the corresponding specimen region. The central challenge is therefore not only to reconstruct each local phase patch accurately, but also to ensure that independently predicted patches remain spatially consistent when assembled into a global phase representation. 

Given a specimen, a 4D-STEM data sample can be represented as 
$(\Phi^{\mathrm{ePIE}}, 
x_n | n=1..N
)$, where
$\Phi^{\mathrm{ePIE}}\in\mathbb{R}^{H_g\times W_g}$ denotes the global phase map, which is reconstructed by ePIE from the set of diffraction measurements 
$\{ x_n | n=1..N \}$. 
$x_n\in\mathbb{R}^{H\times W\times C}$ represents the local diffraction-intensity map acquired at the $n$-th raster-scanned position, where $C$ denotes the number of channels and $H \times W < {H_g\times W_g}$.
The objective is to learn a feed-forward mapping $f_{\theta}$ for the local intensity maps,
\begin{equation}
    \hat{\phi}_n=f_{\theta}(x_n)
    ,
    \label{eq:forward_mapping}
\end{equation}
where $\theta$ denotes the trainable parameters and $\hat{\phi}_n$ is the predicted local phase patch.
The learning objective is to minimize the discrepancy between $\hat{\phi}_n$ and the reference phase patch $\phi_n$, where $\phi_n$ is cropped from the global ePIE reference phase map $\Phi^{\mathrm{ePIE}}$ at the corresponding location.

In the above formulation, the key challenge lies in the local-to-global mapping. Since neighboring scan positions illuminate overlapping specimen regions, their diffraction measurements contain substantial spatial redundancy. Consequently, the independently predicted local phase patches $\hat{\phi}_n$, $n=1,\ldots,N$, must be assembled into a globally consistent phase map. During training, the global ePIE reference phase map can be decomposed into spatially aligned local patches to provide supervision. During inference, however, the model must predict all local phase patches independently from the diffraction measurements and subsequently assemble them into a globally consistent reconstruction.

Our method addresses these constraints by making only local predictions at each scan position, followed by a local-to-global aggregation process to reconstruct the final full-field phase map without requiring an explicit full-field prediction loss or simultaneous processing of the complete 4D-STEM data cube.

\subsection{Data Preprocessing and Augmentation}

The measurable diffraction region at each scan position may exceed the selected input size of $H\times W$. However, under our acquisition conditions, the peripheral regions generally contain sparse or low-signal-to-noise-ratio signals, which may contribute little to phase prediction and may even degrade performance. Therefore, a centered $H\times W$ region is cropped from each diffraction pattern. Since only the diffraction intensity is used, each input is represented as $\mathbf{x}_n\in\mathbb{R}^{H\times W\times1}$. To reduce the large dynamic range of the measured diffraction intensities while preserving sensitivity to weaker signals, element-wise logarithmic compression is applied as
\begin{equation}
\mathbf{x}_{\log,n}=\log(1+\mathbf{x}_n).
\end{equation}
The resulting data are then normalized using statistics estimated exclusively from the source-domain training data. The same normalization is subsequently applied to the target-domain data during evaluation.

In addition, during training, data augmentation techniques are applied to reduce overfitting to dataset-specific intensity distributions. The augmentation includes Poisson noise, additive Gaussian noise, and discrete right-angle rotations. Poisson noise approximates intensity-dependent counting variations, whereas Gaussian noise introduces additional measurement perturbations. For rotational augmentation, an angle is randomly selected from $\{0^\circ,90^\circ,180^\circ,270^\circ\}$ and applied consistently to both the diffraction input and its corresponding phase target. Because only multiples of $90^\circ$ are used, the transformation requires no interpolation. These augmentations expose the model to plausible measurement variations and are intended to reduce overfitting and improve robustness \cite{ongie2020}.

\subsection{Network Architecture}

The proposed model is a phase-only residual U-Net-style encoder--decoder with an additional global context branch, mapping the single-channel diffraction input $\mathbf{x}_n \in \mathbb{R}^{H \times W \times 1}$ at the $n$-th scan position to a two-channel sine--cosine prediction $\hat{\mathbf{y}}_n \in \mathbb{R}^{H_o \times W_o \times 2}$.
Unless otherwise specified, we use $H=W=32$ and $H_o=W_o=16$ as the default configuration.
The encoder consists of three residual stages with 64, 128, and 256 feature channels, respectively, each followed by $2 \times 2$ max pooling. A 512-channel residual bottleneck is used at the lowest $4 \times 4$ spatial resolution. The convolutional blocks use $3 \times 3$ kernels and LeakyReLU activations with a negative slope of 0.15. Spatial dropout with a rate of 0.1 is applied in the deeper encoder stages, bottleneck, decoder, and global branch. U-Net-style encoder--decoder networks provide a suitable balance between contextual feature extraction and spatially resolved prediction~\cite{ronneberger2015}, while residual connections facilitate the optimization of deeper feature transformations~\cite{he2016}.

Inspired by the spatial Fourier transform (SPAF) module introduced in Fourier Imager Network (FIN) \cite{chen2022}, we employ unit-modulus Fourier modulation (UFM) blocks in the encoder, bottleneck, and global context branch. Unlike the original FIN SPAF module, which applies a learned frequency-domain linear transformation within truncated spectral windows, our block modulates the Fourier-domain feature representation using a learnable unit-modulus phasor shared across feature channels. Given an intermediate feature map $\mathbf{F}$, the block first computes its two-dimensional Fourier transform. The resulting frequency-domain representation is multiplied element-wise by a learnable unit-modulus phasor
\begin{equation}
    P(u,v) = \exp\left(i\psi(u,v)\right),
\end{equation}
where $\psi(u,v)\in\mathbb{R}$ denotes the learnable frequency-domain phase modulation and $i$ is the imaginary unit. Consequently, $|P(u,v)|=1$ at every frequency coordinate $(u,v)$, so the modulation changes the phase of each Fourier coefficient without directly scaling its magnitude.

At the bottleneck, an atrous spatial pyramid pooling (ASPP) module aggregates multi-scale context using a $1\times1$ convolution, parallel $3\times3$ convolutions with dilation rates of $1$, $2$, and $3$, and an image-level pooling branch \cite{chen2018}. The bottleneck representation is then progressively restored through two bilinear upsampling stages to the output resolution $H_o\times W_o=16\times16$. The phase-target size is treated as a tunable design choice, and different target sizes are evaluated in the ablation study.

In parallel with the encoder--decoder pathway, a 256-channel global Fourier-inspired branch processes the input $\mathbf{x}_n$ using three consecutive UFM blocks. At the final prediction stage, features from the three encoder levels, the decoded bottleneck representation, and the global branch are resized to the output resolution and fused using a convolutional block attention module (CBAM)~\cite{woo2018}. The fused representation is refined by 128- and 64-channel convolutional residual blocks and projected through a $1 \times 1$ convolution to produce a pre-normalized two-channel representation $\mathbf{z}_n \in \mathbb{R}^{H_o \times W_o \times 2}$. 

\begin{figure*}[t]
    \centering
    \includegraphics[width=0.98\textwidth]{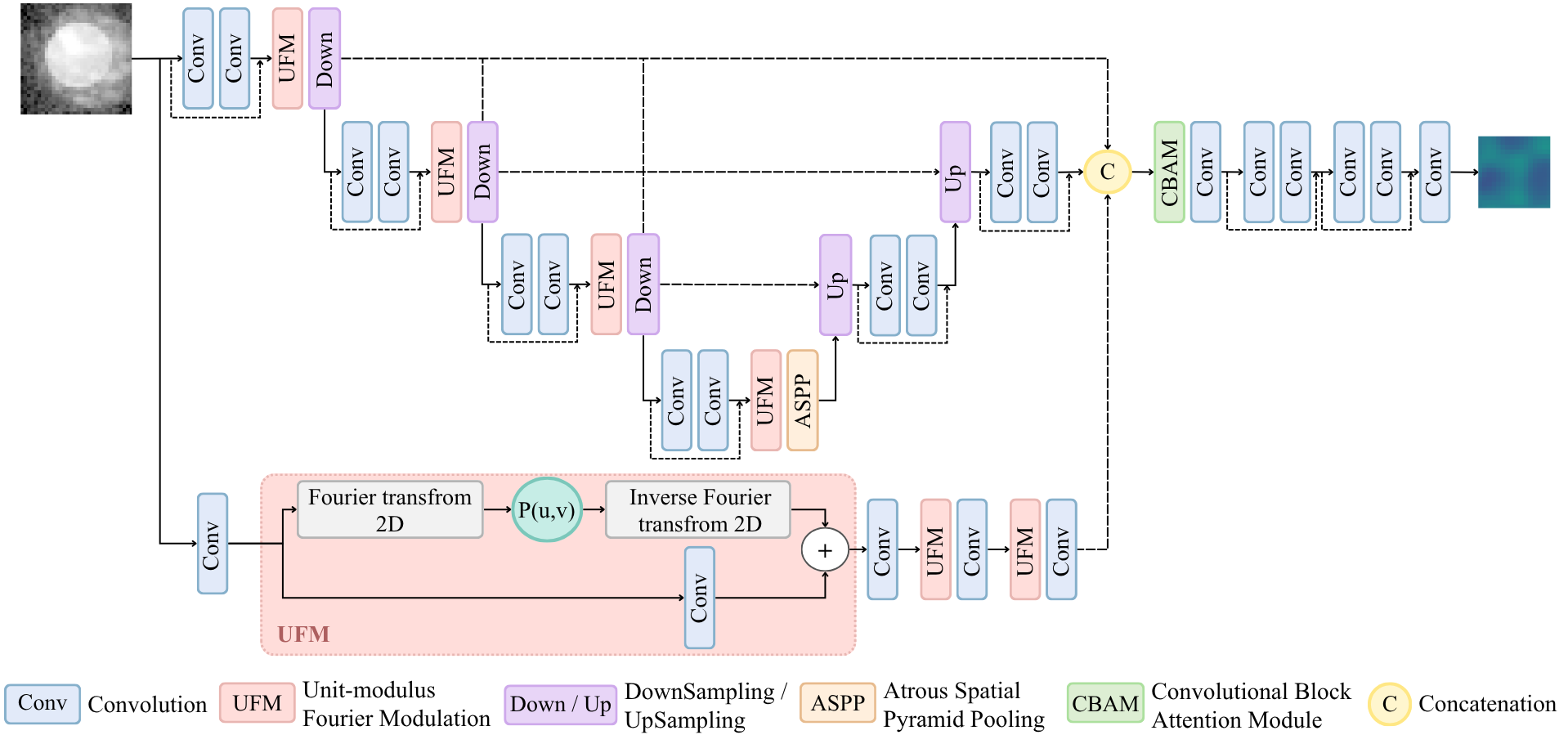}
    \caption{Architecture of the proposed residual U-Net with UFM, ASPP, a global context branch, and CBAM-based feature fusion.}
    \label{fig:architecture}
\end{figure*}

\subsection{Phase Parameterization}

The output at each spatial location of $\mathbf{z}_n$ is a two-dimensional vector, whereas the reference value at each location of $\phi_n$ is a phase in the range $[-\pi,\pi)$.
Direct phase regression is challenging because the wrapped phase is periodic. 
In particular, values near $-\pi$ and $\pi$ represent physically similar phases but are numerically far apart. 
We therefore represent each phase value in $\phi_n$ as a point on the unit circle using its sine and cosine components, thereby preserving angular continuity and avoiding the phase discontinuity at the wrapping boundary \cite{mardia2000,cremers2018,liu2026}.

Specifically, let $\phi_n \in [-\pi,\pi)^{H_o \times W_o}$ be the reference phase patch associated with the $n$-th scan position. 
We represent each reference phase value by its sine and cosine components to obtain
\begin{equation}
    \mathbf{y}_n = [\sin(\phi_n),\cos(\phi_n)]
    \in \mathbb{R}^{H_o \times W_o \times 2}
    .
\end{equation}

Let $\mathbf{z}_n \in \mathbb{R}^{H_o \times W_o \times 2}$ denote the output of the aforementioned phase reconstruction U-Net given the input $\mathbf{x}_n$.
The two-dimensional vector at each spatial location $(i,j)$ of $\mathbf{z}_n$ is normalized to unit length:
\begin{equation}
    \hat{\mathbf{y}}_{n} (i, j)
    =
    \frac{\mathbf{z}_{n} (i, j)}
    {\sqrt{\max\left(\lVert \mathbf{z}_{n} (i, j) \rVert_2^2,\epsilon\right)}},
\end{equation}
where $\lVert\cdot\rVert_2$ denotes the Euclidean norm and $\epsilon>0$ is a small constant for numerical stability. 
The normalized vector $\hat{\mathbf{y}}_{n}(i,j)$ naturally yields the predicted phase:
\begin{equation}
    \hat{\phi}_{n} (i, j)
    =
    \operatorname{atan2}
    \left(
    \hat{\mathbf{y}}^{(1)}_{n} (i, j)
    ,
    \hat{\mathbf{y}}^{(2)}_{n} (i, j)
    \right)
    \in
    [-\pi,\pi)
    ,
\end{equation}
where superscripts $(1)$ and $(2)$ denote the first and second vector components, respectively.

\subsection{Training Objective}

The training objective combines a sine--cosine vector regression loss and an SSIM-based structural loss. Following Section~3.4, let $\mathbf{y}_n$ and $\hat{\mathbf{y}}_n$ denote the reference and predicted sine and cosine representations for the $n$-th scan position, respectively. For notational simplicity, the sample index $n$ is omitted in the following loss expressions.

The vector regression loss is defined as
\begin{equation}
\mathcal{L}_{\mathrm{vec}}
=
\operatorname{mean}
\left(
\left\|\mathbf{y}-\hat{\mathbf{y}}\right\|_2^2
\right),
\end{equation}
where $\operatorname{mean}(\cdot)$ denotes averaging over the training samples and spatial locations, and $\|\cdot\|_2$ denotes the Euclidean norm of the two-dimensional sine--cosine vector.

To encourage preservation of local spatial structures, an SSIM-based loss is also applied to the predicted sine and cosine channels:
\begin{equation}
\mathcal{L}_{\mathrm{SSIM}}
=
1-\frac{1}{2}
\left[
\operatorname{SSIM}(\mathbf{y}_{\sin},\hat{\mathbf{y}}_{\sin})
+
\operatorname{SSIM}(\mathbf{y}_{\cos},\hat{\mathbf{y}}_{\cos})
\right],
\end{equation}
where $\mathbf{y}_{\sin}$ and $\mathbf{y}_{\cos}$ denote the reference sine and cosine channels, while $\hat{\mathbf{y}}_{\sin}$ and $\hat{\mathbf{y}}_{\cos}$ denote their corresponding predictions.

The final phase loss is
\begin{equation}
\mathcal{L}_{\mathrm{phase}}
=
\alpha\mathcal{L}_{\mathrm{vec}}
+
(1-\alpha)\mathcal{L}_{\mathrm{SSIM}},
\end{equation}
where $\alpha\in[0,1]$ controls the relative contribution of the vector regression loss and the structural similarity loss. 
Before computing the SSIM-based loss, the reference and predicted sine and cosine channels are transformed from $[-1,1]$ to $[0,1]$.

Unless otherwise specified, the main experiments use the balanced setting $\alpha=0.5$.

\subsection{Gaussian-Window Global Reconstruction}

Our reconstruction model is trained \textit{only} on local patches at individual scan positions. During inference, the full-field reconstruction is obtained by directly stitching the local predictions using a Gaussian-based algorithm.

First, the predicted local phase patches are placed into the global reconstruction grid according to their calibrated scan positions. Let $r_n=(r_{n,y},r_{n,x})$ denote the top-left object-space coordinate of the phase patch associated with the $n$-th scan position. This coordinate is obtained after applying the crop-offset correction, aligning the scan-coordinate system with the ePIE reference, and rounding to the nearest integer pixel. The same coordinate convention is used during training to extract the corresponding local supervision target from the aligned ePIE reference.

Each prediction $\hat{\phi}_n$ covers only a partial region of the global object-space. Let $r=(r_y,r_x)$ denote an arbitrary location in the global object space. The set of predicted patches covering $r$ is defined as
\begin{equation}
\mathcal{N}(r)
=
\left\{
n \mid r \text{ lies within the spatial support of } \hat{\phi}_n
\right\}.
\end{equation}
For each $n\in\mathcal{N}(r)$, let $p_n(r)=r-r_n$ denote the corresponding local coordinate within the $n$-th patch.

In practice, the overlap among prediction patches is substantial, resulting in a large $|\mathcal{N}(r)|$. Our initial experiments reveal \textit{patch-boundary artifacts} when naive averaging is used. We thus use a Gaussian window to assign larger weights near a local patch center and smaller weights near its boundary. Let $c=(c_y,c_x)$ denote the geometric center of the local patch, where $c_y=(H_o-1)/2$ and $c_x=(W_o-1)/2$. For a local coordinate $p=(p_y,p_x)$, its Gaussian weight is
\begin{equation}
w(p)
=
\exp\left(
-\frac{(p_y-c_y)^2+(p_x-c_x)^2}{2\sigma^2}
\right),
\end{equation}
where $\sigma$ controls the width of the Gaussian window. Unless otherwise specified, we use $\sigma=H_o/6$, with $H_o=W_o=16$ (refer to the ablation study).

Because wrapped phase is periodic, overlapping predictions are combined in the complex plane through weighted circular averaging. For each location $r$ such that $\mathcal{N}(r)\neq\emptyset$, the stitched complex representation is
\begin{equation}
q(r)
=
\frac{
\displaystyle\sum_{n\in\mathcal{N}(r)}
w\!\left(p_n(r)\right)
\exp\left(
i\hat{\phi}_n\!\left(p_n(r)\right)
\right)
}{
\displaystyle\sum_{n\in\mathcal{N}(r)}
w\!\left(p_n(r)\right)
},
\end{equation}
where $q(r)\in\mathbb{C}$ denotes the weighted circular average and $i$ is the imaginary unit.

Following stitching, a light mask-normalized Gaussian smoothing operator $\mathcal{S}_{\rho}$ is applied separately to the real and imaginary components of $q$. The final complex reconstruction is
\begin{equation}
q_{\mathrm{final}}(r)
=
(1-\beta)q(r)
+
\beta\mathcal{S}_{\rho}(q)(r),
\end{equation}
where $\rho$ denotes the standard deviation of the Gaussian smoothing kernel and $\beta\in[0,1]$ controls the contribution of the smoothed result. The reported configuration uses $\rho=1.0$ pixel and $\beta=0.25$.

The final full-field wrapped phase is obtained as
\begin{equation}
\hat{\Phi}(r)
=
\arg\left(q_{\mathrm{final}}(r)\right),
\end{equation}
where $\arg(\cdot)$ returns the complex phase angle in $[-\pi,\pi)$. The reconstruction is defined for object-space locations covered by at least one predicted patch. This procedure reduces boundary discontinuities while respecting the periodic nature of wrapped phase and does not require iterative object--probe updates during inference.

\section{Experiments}
\subsection{Datasets and Evaluation Protocol}
\label{sec:datasets}

For the main experiments, we use four independently acquired 4D-STEM datasets, comprising two AuPd datasets and two MoS$_2$ datasets. One dataset of each material is designated as the source dataset and the other as the target dataset. The cross-material experiments pair the source dataset of one material with the target dataset of the other, whereas the intra-material experiments use independently acquired source and target datasets of the same material. For each source dataset, a fixed set of 15,000 scan positions is randomly sampled for model training. Additional zero-shot validation on an unseen HEA-NP dataset is provided in Appendix~A.

In our datasets, each scan position records a wide $63 \times 63$ diffraction-intensity pattern. A centered $H \times W = 32 \times 32$ region is cropped because most of the measured diffraction intensity is concentrated near the detector center, whereas the peripheral region generally contains weaker signals. This empirically selected crop retains the informative central diffraction structure and is specific to the present acquisition setting.

The probe positions are separated by a scan step of $0.5$~\AA, corresponding to approximately $2.58$ pixels in the reconstructed object space. Each AuPd dataset contains a $400 \times 400$ scan grid with 160,000 diffraction patterns, whereas each MoS$_2$ dataset contains a $300 \times 300$ scan grid with 90,000 diffraction patterns. The corresponding ePIE reference phase maps have spatial dimensions of $1023 \times 1023$ and $583 \times 583$ pixels, respectively. Table~\ref{tab:dataset_summary} summarizes the dataset characteristics.

\begin{table}[t]
\centering
\caption{Summary of the main benchmark dataset characteristics. Each material includes two independently acquired datasets. The scan grid, number of diffraction patterns, and reference object size are reported for each individual dataset.}
\label{tab:dataset_summary}
\resizebox{\columnwidth}{!}{
\begin{tabular}{lcccc}
\toprule
Material & No. of datasets & Scan grid per dataset & Patterns per dataset & Reference object size \\
\midrule
Au--Pd alloy & 2 & $400 \times 400$ & 160,000 & $1023 \times 1023$ \\
MoS$_2$ & 2 & $300 \times 300$ & 90,000 & $583 \times 583$ \\
\bottomrule
\end{tabular}
}
\end{table}

The ePIE reference phase maps are generated using six epochs of blind mixed-state ePIE. The diffraction measurements are treated as amplitudes with centered Fourier transforms, and the relative scan coordinates are interpreted in $y$--$x$ order as top-left probe positions. The object and probe update coefficients are set to $\alpha_{\mathrm{ePIE}}=0.5$ and $\beta_{\mathrm{ePIE}}=0.01$, respectively. Two mutually incoherent probe modes are initialized from the average diffraction signal of 2,048 frames; probe updates begin at the second epoch, with circular probe support and mode orthogonalization applied after each epoch. No stored reference object or probe values are used during the iterative reconstruction.

The primary evaluation setting is cross-material zero-shot inference, in which the model is trained on one material and evaluated directly on the other without fine-tuning. The target-domain datasets are used only for evaluation and are not used for model training or fine-tuning. We consider both AuPd $\rightarrow$ MoS$_2$ and MoS$_2$ $\rightarrow$ AuPd directions. We additionally evaluate the intra-material AuPd $\rightarrow$ AuPd and MoS$_2$ $\rightarrow$ MoS$_2$ settings, where the source and target are independently acquired datasets of the same material.

Before quantitative evaluation, small residual translational offsets between the reconstructed and reference phase maps are corrected using subpixel registration based on cross-correlation~\cite{guizar2008}. We report MSE, PSNR, and MS-SSIM over the selected regions of interest and the valid full-field region defined by the reconstruction mask.

\subsection{Cross-Material Benchmarking}
\label{sec:cross_material}

In both AuPd $\rightarrow$ MoS$_2$ and MoS$_2$ $\rightarrow$ AuPd transfer directions, all learning-based methods are trained using the same fixed source-domain training subset and are evaluated on all scan positions of the target dataset without fine-tuning. The same preprocessing, coordinate calibration, valid reconstruction mask, and evaluation protocol are applied to all methods. We compare the proposed method with PtychoNN and PPN. Figs. 3 and 4 qualitatively compare the full-field reconstruction results and representative regions of interest, while Table 2 presents the corresponding quantitative results.

\begin{table*}[t]
\caption{Full-field reconstruction results for cross-material and intra-material transfer. 
\textbf{Boldface} indicates the best result in each setting.
}
\label{tab:fullfield_benchmark}
\centering
\footnotesize
\setlength{\tabcolsep}{5pt}
\begin{tabular}{lllccc}
\toprule
Evaluation setting & Training $\rightarrow$ testing & Method & MSE $\downarrow$ & PSNR (dB) $\uparrow$ & MS-SSIM $\uparrow$ \\
\midrule
\multirow{6}{*}{Cross-material} & \multirow{3}{*}{AuPd $\rightarrow$ MoS$_2$} & Ours & \textbf{0.007819} & \textbf{21.24} & \textbf{0.6757} \\
& & PtychoNN & 0.009865 & 20.27 & 0.5252 \\
& & PPN & 0.015426 & 18.32 & 0.1150 \\
\cmidrule(lr){2-6}
& \multirow{3}{*}{MoS$_2$ $\rightarrow$ AuPd} & Ours & \textbf{0.005863} & \textbf{24.52} & \textbf{0.6573} \\
& & PtychoNN & 0.006916 & 23.96 &  0.5616 \\
& & PPN & 0.006042 & 24.42 & 0.5891 \\
\midrule
\multirow{6}{*}{Intra-material} & \multirow{3}{*}{AuPd $\rightarrow$ AuPd} & Ours & \textbf{0.002990} & \textbf{29.69} & \textbf{0.7295} \\
& & PtychoNN & 0.003473 & 29.41 & 0.7249 \\
& & PPN & 0.003843 & 28.87 & 0.6150 \\
\cmidrule(lr){2-6}
& \multirow{3}{*}{MoS$_2$ $\rightarrow$ MoS$_2$} & Ours & 0.005261 & 22.95 & \textbf{0.8311} \\
& & PtychoNN & \textbf{0.005084} & \textbf{23.14} & 0.8194 \\
& & PPN & 0.006561 & 21.99 & 0.7538 \\
\bottomrule
\end{tabular}
\end{table*}

The AuPd reference contains both crystalline and amorphous regions. Consequently, its full-field metrics reflect reconstruction performance over both periodic lattice structures and spatially irregular amorphous areas. We therefore interpret the full-field measurements together with the crystalline ROIs when assessing structural preservation.

\begin{figure*}[t]
    \centering
    \includegraphics[width=0.98\textwidth]{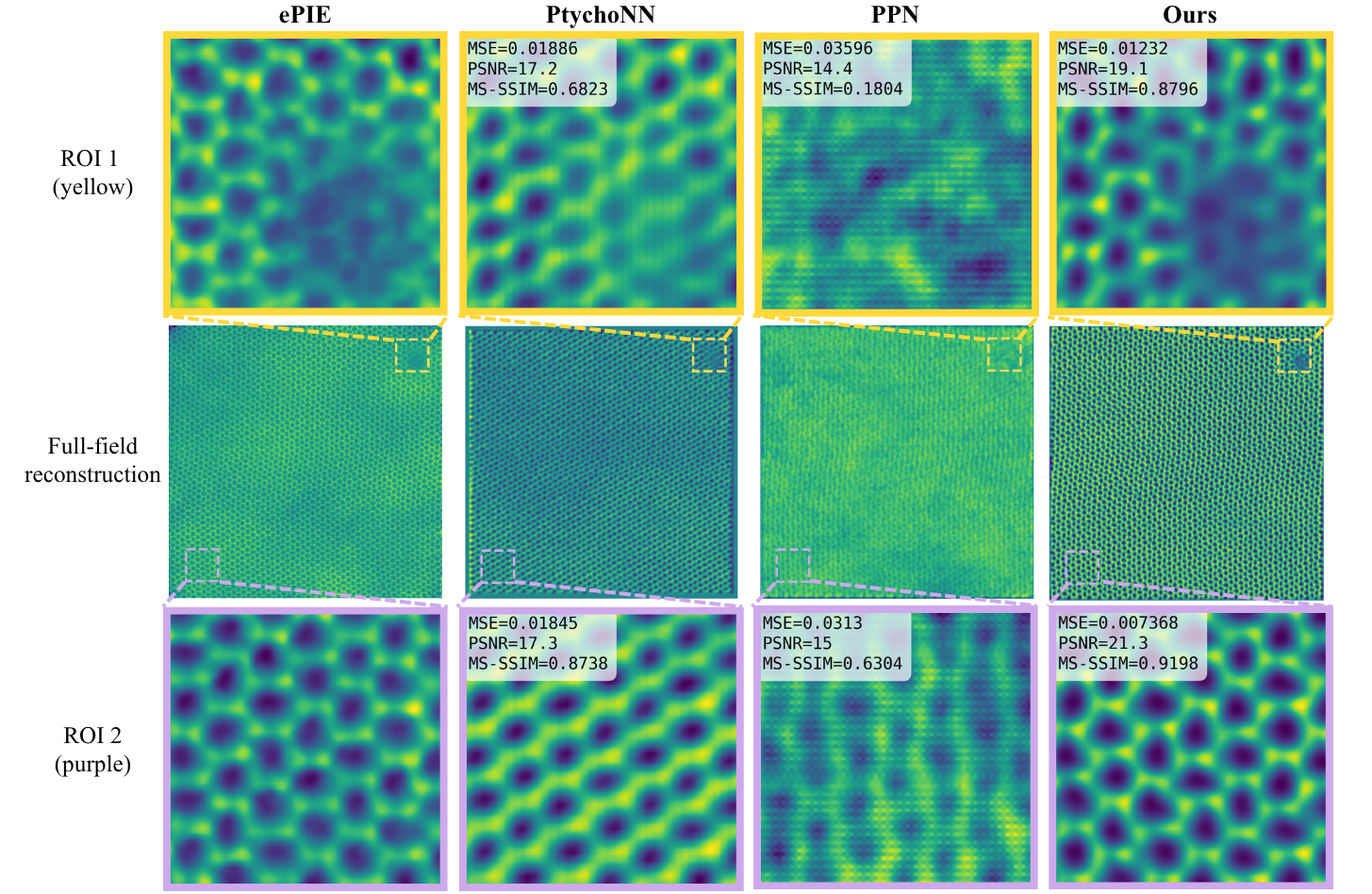}
    \caption{Zero-shot cross-material reconstruction for AuPd $\rightarrow$ MoS$_2$. 
    The ePIE column shows the reference reconstruction. We compare with PtychoNN and PPN. 
    The middle row shows the full-field reconstructions, while the top and bottom rows present two representative ROIs.
    The overlaid metrics are computed over the corresponding ROIs.
    }
    \label{fig:cross_aupd_mos2}
\end{figure*}

For AuPd $\rightarrow$ MoS$_2$, the proposed method preserves the periodic phase morphology more consistently, whereas PtychoNN exhibits directional smearing and PPN introduces pronounced stripe artifacts. The proposed method achieves the best full-field MSE, PSNR, and MS-SSIM. The ROI results show the same general trend, although PtychoNN obtains a slightly higher MS-SSIM in the lower ROI.

\begin{figure*}[t]
    \centering
    \includegraphics[width=0.98\textwidth]{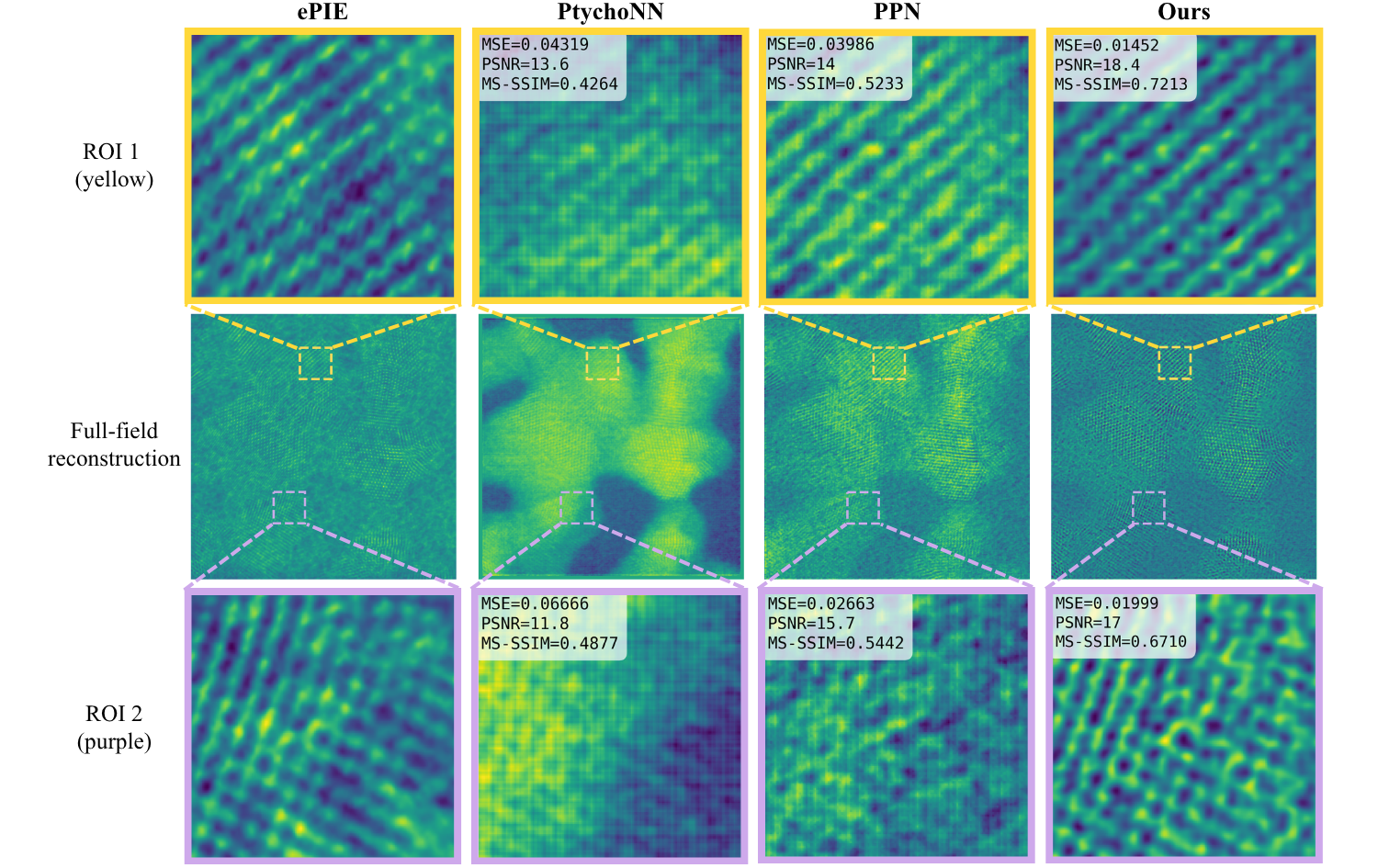}
    \caption{Zero-shot cross-material reconstruction for MoS$_2$ $\rightarrow$ AuPd. 
    The middle row shows the full-field reconstructions, while the top and bottom rows present two representative ROIs.
    }
    \label{fig:cross_mos2_aupd}
\end{figure*}

For MoS$_2$ $\rightarrow$ AuPd, PtychoNN exhibits strong low-frequency variations and loses fine local structures, while PPN retains more detail but produces a noisier phase texture. PPN obtains better pixel-wise metrics in the upper ROI, whereas the proposed method achieves substantially higher structural similarity. The proposed method performs best across all three metrics in the lower ROI and in the full-field evaluation.

Overall, the proposed method achieves the best full-field performance in both cross-material directions, indicating more consistent reconstruction quality under zero-shot transfer. The localized ROI differences further suggest that pixel-wise accuracy and multi-scale structural similarity may favor different methods in individual regions.

\subsection{Intra-Material Benchmarking}
\label{sec:intra_material}

\begin{figure*}[t]
    \centering
    \includegraphics[width=0.98\textwidth]{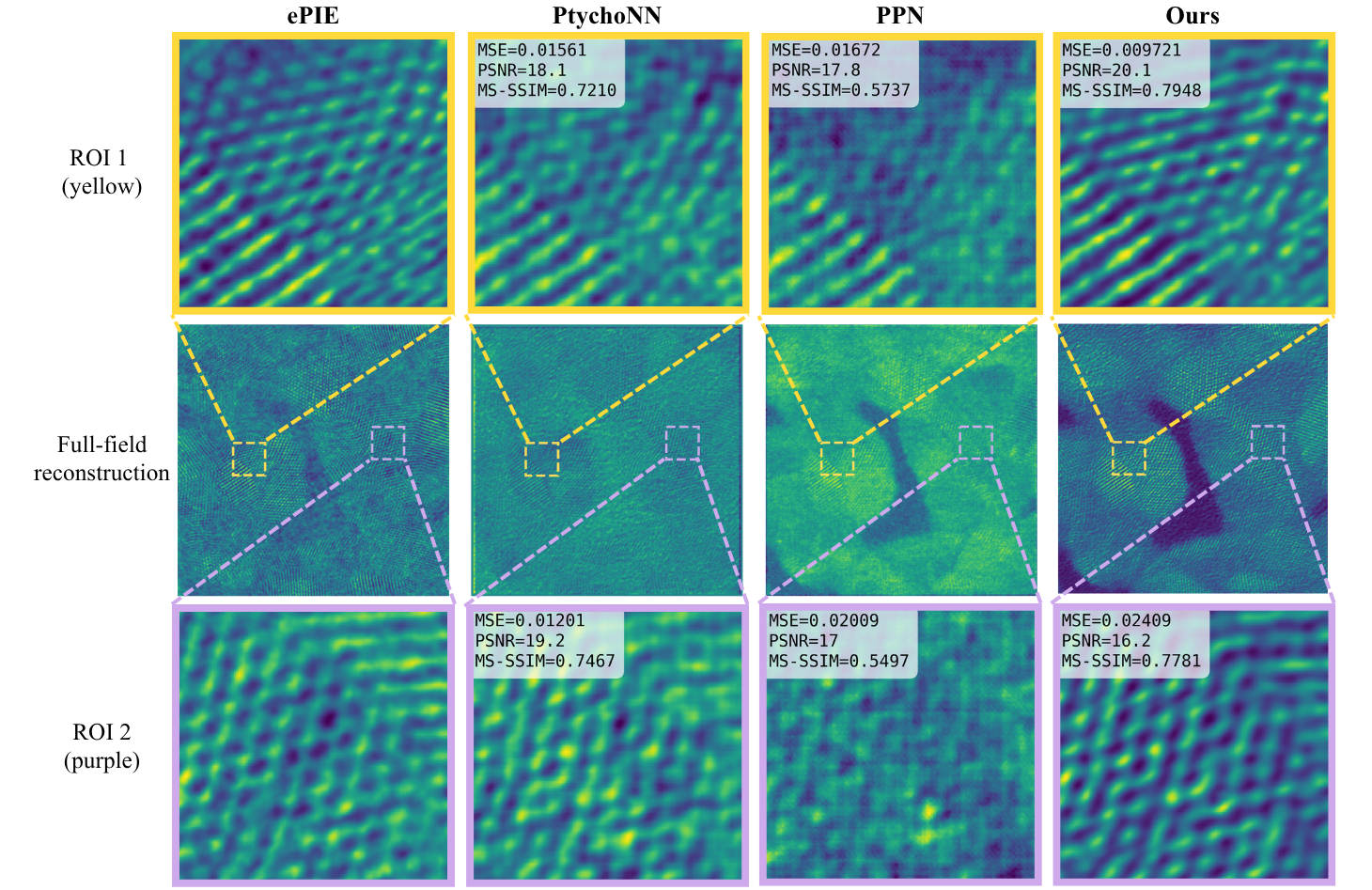}
    \caption{Intra-material reconstruction for AuPd $\rightarrow$ AuPd.  
    The middle row shows the full-field reconstructions, while the top and bottom rows present two representative ROIs.
    }
    \label{fig:intra_aupd}
\end{figure*}

\begin{figure*}[t]
    \centering
    \includegraphics[width=0.98\textwidth]{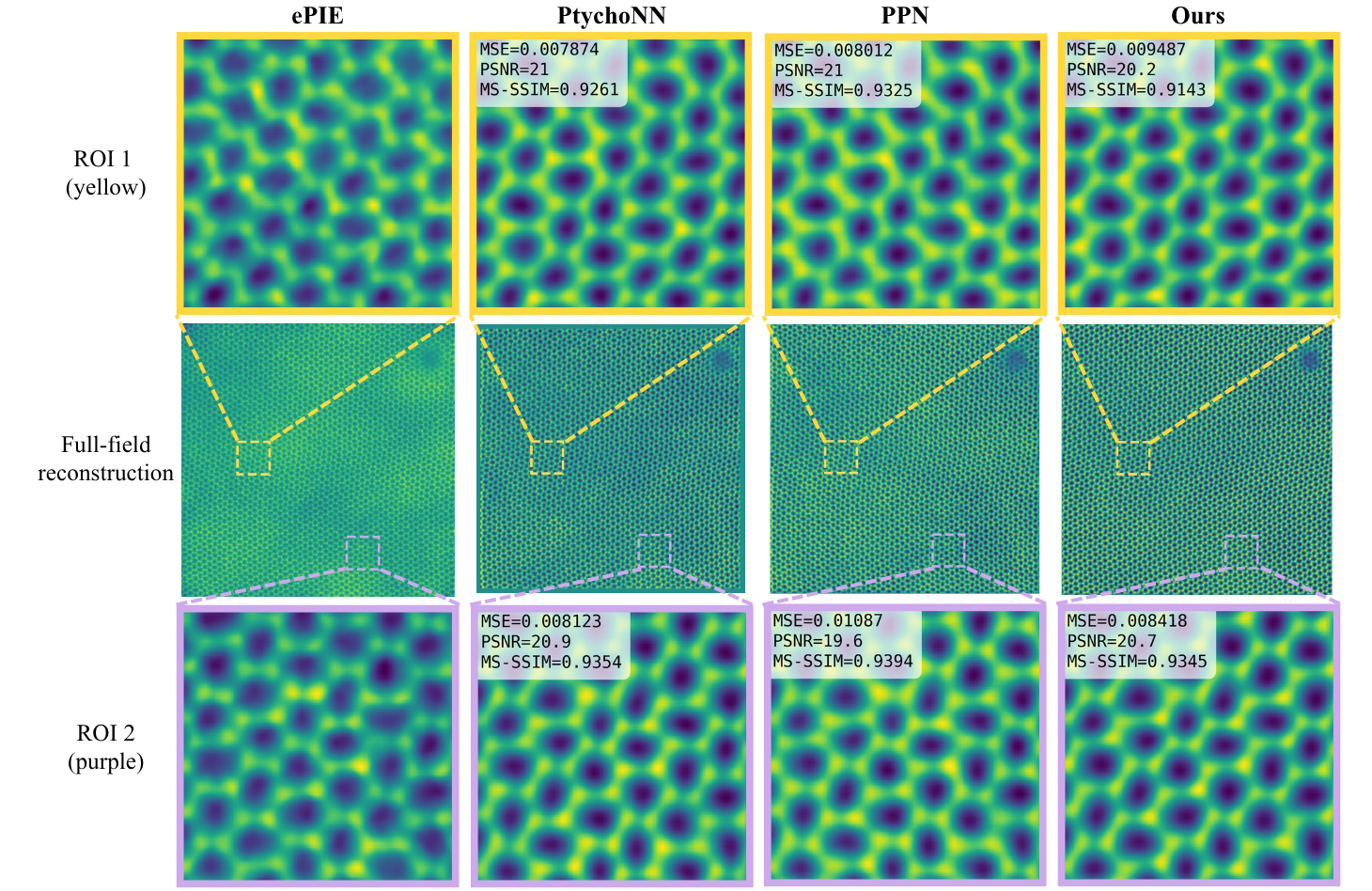}
    \caption{Intra-material reconstruction for MoS$_2$ $\rightarrow$ MoS$_2$. 
    The middle row shows the full-field reconstructions, while the top and bottom rows present two representative ROIs.
    }
    \label{fig:intra_mos2}
\end{figure*}

The intra-material AuPd $\rightarrow$ AuPd and MoS$_2$ $\rightarrow$ MoS$_2$ settings provide a lower-domain-shift counterpart to the cross-material evaluation. 
Quantitative results are reported in Table~\ref{tab:fullfield_benchmark}, while qualitative results are presented in Figs.~\ref{fig:intra_aupd} and~\ref{fig:intra_mos2}.


For AuPd $\rightarrow$ AuPd, the proposed method preserves the directional phase structures in the upper ROI and achieves the best result across all three ROI metrics. In the lower ROI, PtychoNN obtains lower MSE and higher PSNR, whereas the proposed method achieves a slightly higher MS-SSIM. The proposed method also achieves the best full-field performance across all three metrics. 
The AuPd full-field evaluation includes both crystalline and amorphous regions and is therefore interpreted together with the selected crystalline ROIs.

For MoS$_2$ $\rightarrow$ MoS$_2$, all three learning-based methods preserve the periodic lattice orientation. PPN performs best in the upper ROI, while PtychoNN performs best in the lower ROI; the proposed method nevertheless retains comparable periodic phase morphology. In the full-field evaluation, PtychoNN achieves the lowest MSE and highest PSNR, whereas the proposed method achieves the highest MS-SSIM, reflecting a trade-off between pixel-wise accuracy and multi-scale structural similarity.

Overall, the proposed method achieves the best full-field performance for AuPd $\rightarrow$ AuPd and the highest full-field MS-SSIM for MoS$_2$ $\rightarrow$ MoS$_2$. These results indicate stable structural preservation across independently acquired datasets of the same material, although the best pixel-wise accuracy depends on the specific source--target dataset pair.

\subsection{Runtime Analysis}
\label{sec:runtime}

We compare the end-to-end reconstruction time of ePIE and the proposed pipeline on the AuPd and MoS$_2$ datasets. Both methods process the same number of scan positions and produce final phase maps with identical spatial dimensions. However, ePIE repeatedly revisits the diffraction patterns from all scan positions over multiple iterative passes, whereas the proposed pipeline processes each scan position once through feed-forward inference and subsequently assembles the predicted local phase patches into a full-field reconstruction.

Both methods were evaluated using the same NVIDIA Tesla V100-SXM2 GPU with 16~GB of GPU memory. All timing values are reported as the mean and standard deviation over five runs. For ePIE, the total runtime covers the complete iterative reconstruction procedure required to obtain the final object phase map. For the proposed pipeline, the total runtime includes preprocessing, batching, data transfer, GPU synchronization, neural-network inference, and full-field Gaussian-window stitching. The per-frame values report the average processing time per scan position for each stage of the proposed pipeline. The end-to-end speedup is computed by dividing the mean total runtime of ePIE by the mean total runtime of the proposed pipeline.

\begin{table*}[t]
\centering
\caption{End-to-end runtime comparison between ePIE and the proposed pipeline using the same NVIDIA Tesla V100-SXM2 GPU. Values are reported as mean $\pm$ standard deviation over five runs, with per-frame processing times measured in milliseconds. The speedup is calculated by dividing the mean ePIE runtime by the mean runtime of the proposed pipeline.}
\label{tab:runtime}
\resizebox{\textwidth}{!}{
\begin{tabular}{llccccccc}
\toprule
Method & Dataset & Recon. size & \# Patterns & Preproc. / frame (ms) & Inference / frame (ms) & Stitching / frame (ms) & Total time (s) & Speedup \\
\midrule
ePIE & AuPd & $1023 \times 1023$ & 160,000 & -- & -- & -- & $941.00 \pm 6.82$ & $1.00\times$ \\
Proposed & AuPd & $1023 \times 1023$ & 160,000 & $0.016 \pm 0.002$ & $0.497 \pm 0.008$ & $0.051 \pm 0.005$ & $94.05 \pm 1.41$ & $\mathbf{10.01\times}$ \\
ePIE & MoS$_2$ & $583 \times 583$ & 90,000 & -- & -- & -- & $542.00 \pm 5.49$ & $1.00\times$ \\
Proposed & MoS$_2$ & $583 \times 583$ & 90,000 & $0.021 \pm 0.003$ & $0.501 \pm 0.011$ & $0.054 \pm 0.004$ & $54.50 \pm 0.95$ & $\mathbf{9.94\times}$ \\
\bottomrule
\end{tabular}
}
\end{table*}

As shown in Table~\ref{tab:runtime}, the proposed pipeline reduces the mean total reconstruction time from $941.00 \pm 6.82$~s to $94.05 \pm 1.41$~s for AuPd and from $542.00 \pm 5.49$~s to $54.50 \pm 0.95$~s for MoS$_2$. These results correspond to end-to-end speedups of $10.01\times$ and $9.94\times$, respectively. Because both methods were evaluated on the same GPU, the reported values provide a direct same-GPU runtime comparison. The relatively small standard deviations indicate limited run-to-run variation under the evaluated configurations.

Neural-network inference is the dominant component of the proposed pipeline, requiring approximately $0.50$~ms per diffraction pattern for both datasets. By comparison, preprocessing and Gaussian-window stitching introduce substantially smaller per-frame costs. These results demonstrate that, after offline training, the proposed feed-forward pipeline reduces online reconstruction latency by approximately one order of magnitude relative to iterative ePIE reconstruction on the same GPU.

\subsection{Ablation Study}
\label{sec:ablation}

We conduct ablation experiments on five design aspects: the network architecture, the supervised phase-target size, the loss-weight coefficient $\alpha$, the Gaussian-window width, and the stitching strategy. Unless otherwise specified, the default configuration uses the full network architecture, a $H_o \times W_o = 16\times16$ phase target, $\alpha=0.5$, and Gaussian-window stitching with $\sigma=H_o/6$. 

\begin{figure*}[t]
    \centering
    \includegraphics[width=0.98\textwidth]{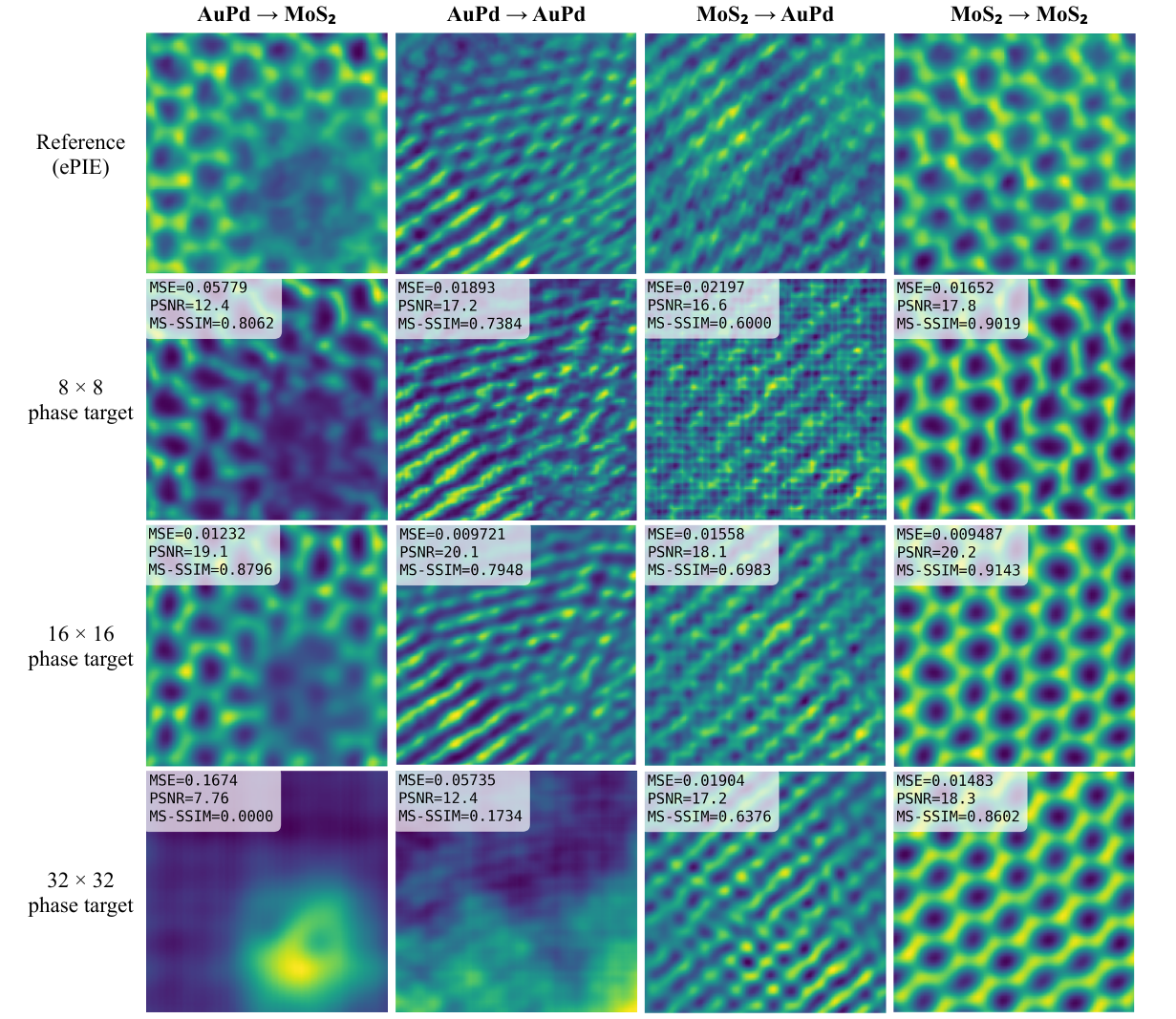}
    \caption{Representative phase-target-size ablations. Columns correspond to AuPd $\rightarrow$ MoS$_2$, AuPd $\rightarrow$ AuPd, MoS$_2$ $\rightarrow$ AuPd, and MoS$_2$ $\rightarrow$ MoS$_2$. Rows show the ePIE references and reconstructions obtained using $8\times8$, $16\times16$, and $32\times32$ targets. The overlaid metrics are ROI-based, and the full-field results are reported in Table~\ref{tab:ablation_all}.}
    \label{fig:target_size_ablation}
\end{figure*}

\begin{figure*}[t]
    \centering
    \includegraphics[width=0.98\textwidth]{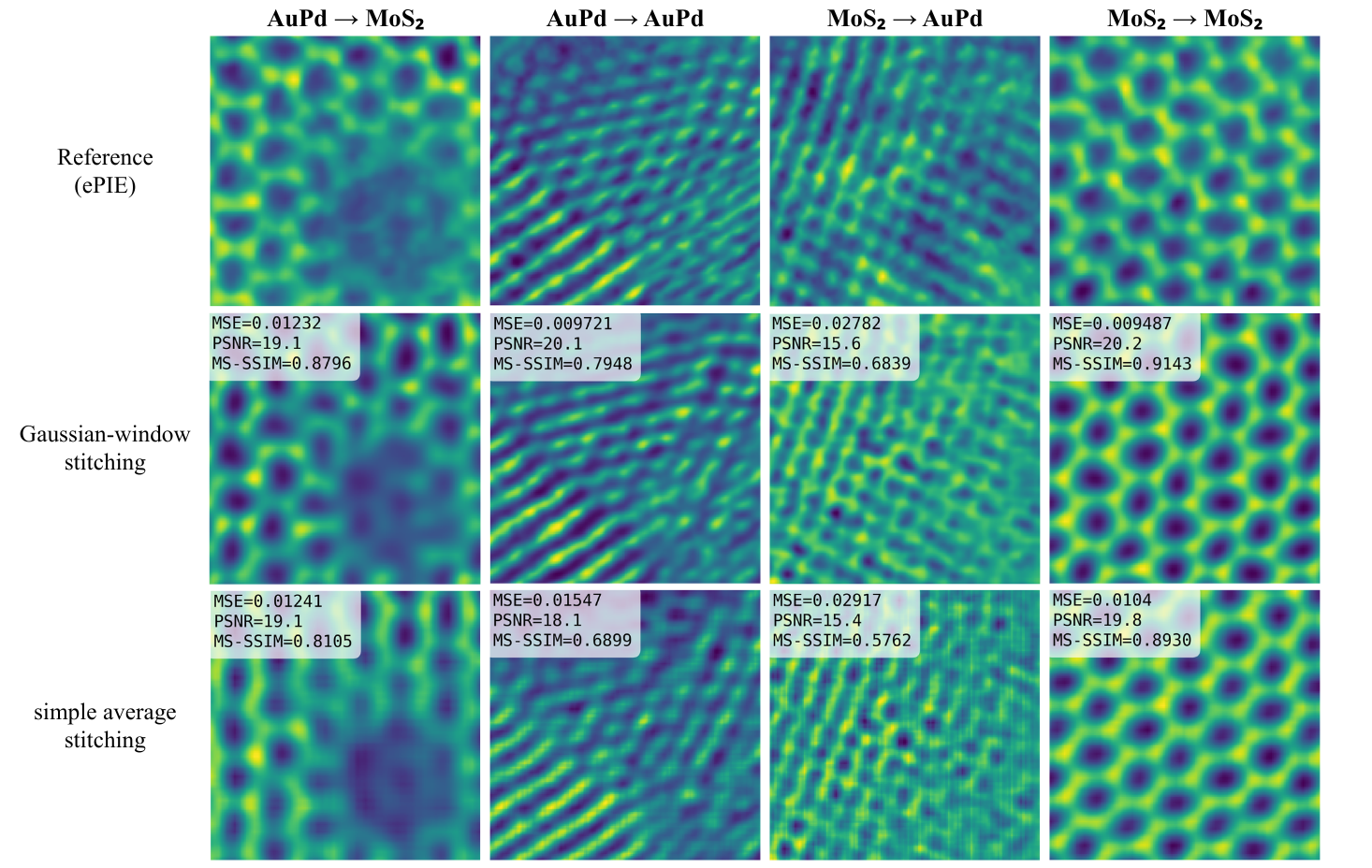}
    \caption{Representative comparison of Gaussian-window and simple-average stitching. Columns correspond to AuPd $\rightarrow$ MoS$_2$, AuPd $\rightarrow$ AuPd, MoS$_2$ $\rightarrow$ AuPd, and MoS$_2$ $\rightarrow$ MoS$_2$. Rows show the ePIE references and the two stitching results. The overlaid metrics are ROI-based, and the full-field results are reported in Table~\ref{tab:ablation_all}.}
    \label{fig:stitching_ablation}
\end{figure*}

All variants are evaluated under the four transfer settings used in Sections~\ref{sec:cross_material} and~\ref{sec:intra_material}. Table~\ref{tab:ablation_all} reports full-field MSE, PSNR, and MS-SSIM using the evaluation protocol described in Section~4.1, while Figures~\ref{fig:target_size_ablation} and~\ref{fig:stitching_ablation} provide representative regional comparisons. The architectural, target-size, and loss-weight variants are trained separately under otherwise identical settings. The Gaussian-width and stitching ablations use predictions from the same default model and modify only the local-to-global aggregation procedure during inference.

\begin{table*}[t]
\caption{Full-field ablation results across four transfer settings. Intra-material pairs use independently acquired source and target datasets. The dagger ($\dagger$) denotes the default configuration, and bold values indicate the best result within each ablation category and transfer setting.}
\label{tab:ablation_all}
\centering
\scriptsize
\resizebox{\textwidth}{!}{%
\begin{tabular}{llcccccccccccc}
\toprule
\multirow{2}{*}{Ablation} &
\multirow{2}{*}{Variant} &
\multicolumn{3}{c}{AuPd $\rightarrow$ MoS$_2$} &
\multicolumn{3}{c}{MoS$_2$ $\rightarrow$ AuPd} &
\multicolumn{3}{c}{AuPd $\rightarrow$ AuPd} &
\multicolumn{3}{c}{MoS$_2$ $\rightarrow$ MoS$_2$} \\
\cmidrule(lr){3-5}
\cmidrule(lr){6-8}
\cmidrule(lr){9-11}
\cmidrule(lr){12-14}
& &
MSE $\downarrow$ & PSNR $\uparrow$ & MS-SSIM $\uparrow$ &
MSE $\downarrow$ & PSNR $\uparrow$ & MS-SSIM $\uparrow$ &
MSE $\downarrow$ & PSNR $\uparrow$ & MS-SSIM $\uparrow$ &
MSE $\downarrow$ & PSNR $\uparrow$ & MS-SSIM $\uparrow$ \\
\midrule

\multirow{6}{*}{Architecture}
& Residual U-Net
& 0.008721 & 20.75 & 0.6155
& 0.006762 & 24.03 & 0.5597
& 0.003144 & 29.43 & 0.7033
& 0.005391 & 22.85 & 0.8229 \\

& w/o UFM
& 0.008205 & 21.01 & 0.6463
& 0.006320 & 24.26 & 0.5894
& 0.003182 & 29.42 & 0.6987
& 0.005366 & 22.88 & 0.8240 \\

& w/o ASPP
& 0.008524 & 20.85 & 0.6272
& 0.006912 & 23.95 & 0.5577
& 0.003271 & 29.40 & 0.6915
& 0.005379 & 22.87 & 0.8245 \\

& w/o Global Branch
& 0.009227 & 20.51 & 0.5765
& 0.006453 & 24.19 & 0.5852
& 0.003260 & 29.36 & 0.6908
& 0.005429 & 22.82 & 0.8223 \\

& w/o CBAM
& 0.015637 & 18.27 & 0.1036
& 0.006387 & 24.20 & 0.5798
& 0.003365 & 29.21 & 0.6814
& 0.005501 & 22.75 & 0.8181 \\

& Full Model\textsuperscript{$\dagger$}
& \textbf{0.007819} & \textbf{21.24} & \textbf{0.6757}
& \textbf{0.005863} & \textbf{24.52} & \textbf{0.6573}
& \textbf{0.002990} & \textbf{29.69} & \textbf{0.7295}
& \textbf{0.005261} & \textbf{22.95} & \textbf{0.8311} \\

\midrule

\multirow{3}{*}{Target size}
& $8\times8$
& 0.008497 & 20.87 & 0.6320
& 0.006293 & 24.28 & 0.5644
& 0.004019 & 28.75 & 0.5965
& \textbf{0.004956} & \textbf{23.18} & \textbf{0.8471} \\

& $16\times16$\textsuperscript{$\dagger$}
& \textbf{0.007819} & \textbf{21.24} & \textbf{0.6757}
& \textbf{0.005863} & \textbf{24.52} & \textbf{0.6573}
& \textbf{0.002990} & \textbf{29.69} & \textbf{0.7295}
& 0.005261 & 22.95 & 0.8311 \\

& $32\times32$
& 0.015609 & 18.28 & 0.1032
& 0.006826 & 24.15 & 0.5213
& 0.004031 & 28.74 & 0.5961
& 0.006420 & 22.12 & 0.7731 \\

\midrule

\multirow{3}{*}{Loss weight}
& $\alpha=0.25$
& 0.007993 & 21.13 & 0.6634
& 0.006300 & 24.27 & 0.5862
& 0.003081 & 29.59 & 0.7143
& 0.005360 & 22.93 & 0.8244 \\

& $\alpha=0.50$\textsuperscript{$\dagger$}
& \textbf{0.007819} & \textbf{21.24} & \textbf{0.6757}
& \textbf{0.005863} & \textbf{24.52} & \textbf{0.6573}
& \textbf{0.002990} & \textbf{29.69} & \textbf{0.7295}
& \textbf{0.005261} & \textbf{22.95} & \textbf{0.8311} \\

& $\alpha=0.75$
& 0.011657 & 19.51 & 0.3976
& 0.006241 & 24.29 & 0.5864
& 0.004014 & 28.75 & 0.5958
& 0.005379 & 22.86 & 0.8230 \\

\midrule

\multirow{3}{*}{Gaussian width}
& $\sigma=H_o/4$
& 0.009082 & 20.58 & 0.5856
& 0.006431 & 24.19 & 0.5830
& 0.003138 & 29.46 & 0.7034
& 0.006055 & 22.35 & 0.7858 \\

& $\sigma=H_o/6$\textsuperscript{$\dagger$}
& \textbf{0.007819} & \textbf{21.24} & \textbf{0.6757}
& \textbf{0.005863} & \textbf{24.52} & \textbf{0.6573}
& \textbf{0.002990} & 29.69 & 0.7295
& 0.005261 & 22.95 & \textbf{0.8311} \\

& $\sigma=H_o/8$
& 0.008287 & 21.05 & 0.6155
& 0.006089 & 24.39 & 0.6010
& 0.003020 & \textbf{29.75} & \textbf{0.7346}
& \textbf{0.004959} & \textbf{23.29} & 0.8026 \\

\midrule

\multirow{2}{*}{Stitching}
& Gaussian\textsuperscript{$\dagger$}
& \textbf{0.007819} & \textbf{21.24} & \textbf{0.6757}
& \textbf{0.005863} & \textbf{24.52} & \textbf{0.6573}
& \textbf{0.002990} & \textbf{29.69} & \textbf{0.7295}
& \textbf{0.005261} & \textbf{22.95} & \textbf{0.8311} \\

& Simple average
& 0.011128 & 19.71 & 0.4386
& 0.006798 & 24.01 & 0.5551
& 0.003445 & 29.16 & 0.6609
& 0.007387 & 21.49 & 0.7026 \\

\bottomrule
\end{tabular}%
}
\end{table*}

\par\smallskip\noindent\textbf{Network architecture.}
We compare a base residual U-Net, four component-removal variants, and the full model. The full model provides the most consistent performance across all four transfer settings, supporting the complementary contributions of UFM, ASPP, the global context branch, and CBAM-based feature fusion. Removing CBAM causes the most pronounced degradation for AuPd $\rightarrow$ MoS$_2$, indicating the importance of attention-based feature fusion under cross-material transfer.

\par\smallskip\noindent\textbf{Target size.}
We compare $8\times8$, $16\times16$, and $32\times32$ phase targets, corresponding to physical widths of approximately $1.55$~\AA, $3.10$~\AA, and $6.20$~\AA, respectively, under the object-space sampling of $0.194$~\AA\ per pixel. The $16\times16$ target provides the most consistent overall performance, while the smaller target is favorable for MoS$_2$ $\rightarrow$ MoS$_2$ and the larger target generally reduces reconstruction quality. The physical width of the $16\times16$ target is close to the in-plane lattice constant of MoS$_2$, approximately $3.16$~\AA~\cite{molina2015vibrational}. It is also comparable to characteristic nearest-neighbor length scales in crystalline face-centered-cubic Au--Pd regions, estimated from reported lattice parameters using the relation $a/\sqrt{2}$~\cite{maeland1964lattice}. This correspondence suggests that the selected target may capture physically relevant local periodic structures. However, because this target-size analysis is conducted only on the two main material systems and the $8\times8$ target performs best for MoS$_2$ $\rightarrow$ MoS$_2$, the $16\times16$ target should be regarded as an empirically effective scale rather than a universally optimal physical scale.

Figure~\ref{fig:target_size_ablation} shows that the $16\times16$ target generally preserves recognizable local phase structures, whereas the smaller and larger targets introduce fragmented or overly smoothed reconstructions in some settings.

\par\smallskip\noindent\textbf{Loss weight.}
We vary $\alpha$ to control the balance between sine--cosine vector regression and SSIM-based structural supervision. The balanced setting $\alpha=0.5$ performs most consistently across the four transfer settings, supporting the joint use of pixel-wise and structural objectives.

\par\smallskip\noindent\textbf{Gaussian-window width.}
We compare $\sigma=H_o/4$, $\sigma=H_o/6$, and $\sigma=H_o/8$. Although the narrower window improves selected metrics in the intra-material settings, $\sigma=H_o/6$ provides the most stable overall balance across transfer directions and is retained as the default setting.

\par\smallskip\noindent\textbf{Stitching strategy.}
Gaussian-window stitching consistently outperforms simple averaging in the full-field evaluation. The center-weighted aggregation reduces the influence of less reliable patch boundaries and produces more spatially coherent reconstructions.

Figure~\ref{fig:stitching_ablation} further shows that simple averaging can introduce visible overlap-related artifacts, whereas Gaussian weighting better preserves the dominant phase morphology.

Overall, the ablation results support the use of the full network architecture, a $16\times16$ phase target, $\alpha=0.5$, and Gaussian-window stitching with $\sigma=H_o/6$ as the default configuration.

\section{Conclusion}

This work presents a \textbf{direct local-to-global} framework for \textbf{zero-shot cross-material} ptychographic reconstruction. The model predicts local wrapped-phase patches from diffraction intensities using a sine--cosine representation and assembles them into full-field phase maps through scan-position-aware Gaussian-window stitching. Without target-domain fine-tuning, the proposed method achieves the best full-field MSE, PSNR, and MS-SSIM among the evaluated learning-based methods in both AuPd $\rightarrow$ MoS$_2$ and MoS$_2$ $\rightarrow$ AuPd transfer directions. Additional zero-shot validation on an unseen HEA-NP dataset further shows that models trained on either MoS$_2$ or AuPd retain substantial reconstruction capability when transferred to a third material domain. The ablation results support the selected network architecture, local-supervision scale, loss formulation, and Gaussian-window stitching strategy. The proposed pipeline also reduces the end-to-end reconstruction time by approximately \textbf{one order} of magnitude compared with iterative ePIE under the evaluated configurations. Overall, these findings demonstrate the potential of local phase prediction as an efficient and transferable approach to cross-material ptychographic reconstruction. While the main benchmark focuses on two material systems and the HEA-NP experiment serves as an additional unseen-domain validation, broader evaluation across more material compositions and acquisition conditions remains necessary. The method also relies on ePIE-derived supervision, so potential reference-reconstruction bias remains to be investigated in future work.

\section*{Acknowledgments}
The authors gratefully acknowledge Prof. Chein-Chun Chen of the Department of Engineering and System Science, National Tsing Hua University, for providing the experimental datasets and the MRDM source code used in this study, thereby enabling the reproduction of the MRDM method and the corresponding comparative experiments. N. Y. Chen thanks the National Center for High-Performance Computing (NCHC), National Institutes of Applied Research (NIAR), Taiwan, for providing computational and storage resources supported by National Science and Technology Council (NSTC), Taiwan, under Grants No. NSTC 114-2112-M-029-008- and NSTC 115-2112-M-029-008-.

{
    \small
    \bibliographystyle{ieeenat_fullname}
    \bibliography{main}

@article{rodenburg2004,
  author  = {Rodenburg, J. M. and Faulkner, H. M. L.},
  title   = {A phase retrieval algorithm for shifting illumination},
  journal = {Applied Physics Letters},
  volume  = {85},
  number  = {20},
  pages   = {4795--4797},
  year    = {2004}
}

@article{maiden2009,
  author  = {Maiden, A. M. and Rodenburg, J. M.},
  title   = {An improved ptychographical phase retrieval algorithm for diffractive imaging},
  journal = {Ultramicroscopy},
  volume  = {109},
  number  = {10},
  pages   = {1256--1262},
  year    = {2009}
}

@article{yu2024,
  author  = {Yu, R. and Sha, H. and Cui, J. and Yang, W.},
  title   = {Introduction to electron ptychography for materials scientists},
  journal = {Microstructures},
  volume  = {4},
  pages   = {2024056},
  year    = {2024}
}

@article{ophus2019,
  author  = {Ophus, C.},
  title   = {Four-dimensional scanning transmission electron microscopy ({4D-STEM}): From scanning nanodiffraction to ptychography and beyond},
  journal = {Microscopy and Microanalysis},
  volume  = {25},
  number  = {3},
  pages   = {563--582},
  year    = {2019}
}

@article{thibault2009,
  author  = {Thibault, P. and Dierolf, M. and Bunk, O. and Menzel, A. and Pfeiffer, F.},
  title   = {Probe retrieval in ptychographic coherent diffractive imaging},
  journal = {Ultramicroscopy},
  volume  = {109},
  number  = {4},
  pages   = {338--343},
  year    = {2009}
}

@article{cherukara2020,
  author  = {Cherukara, M. J. and others},
  title   = {{AI}-enabled high-resolution scanning coherent diffraction imaging},
  journal = {Applied Physics Letters},
  volume  = {117},
  number  = {4},
  pages   = {044103},
  year    = {2020}
}

@article{hoidn2023,
  author  = {Hoidn, O. and Mishra, A. A. and Mehta, A.},
  title   = {Physics constrained unsupervised deep learning for rapid, high resolution scanning coherent diffraction reconstruction},
  journal = {Scientific Reports},
  volume  = {13},
  pages   = {22789},
  year    = {2023}
}

@article{yamada2024,
  author  = {Yamada, K. and Akaishi, N. and Yatabe, K. and Takayama, Y.},
  title   = {Ptychographic phase retrieval via a deep-learning-assisted iterative algorithm},
  journal = {Journal of Applied Crystallography},
  volume  = {57},
  number  = {5},
  pages   = {1323--1335},
  year    = {2024}
}

@misc{nakahata2024,
  author       = {Nakahata, R. and Zaman, S. and Zhang, M. and Lu, F. and Chiu, K.},
  title        = {{PtychoFormer}: A transformer-based model for ptychographic phase retrieval},
  howpublished = {arXiv:2410.17377},
  year         = {2024}
}

@article{yue2025,
  author  = {Yue, H. and Cheng, J. and Ren, Y.-X. and Chen, C.-C. and van Riessen, G. A. and Leong, P. H. W. and Shu, S. F.},
  title   = {A Physics-Inspired Deep Learning Framework With Polar Coordinate Attention for Ptychographic Imaging},
  journal = {IEEE Transactions on Computational Imaging},
  volume  = {11},
  pages   = {888--900},
  year    = {2025},
  doi     = {10.1109/TCI.2025.3572250}
}

@article{chen2022,
  author  = {Chen, H. and Huang, L. and Liu, T. and Ozcan, A.},
  title   = {Fourier Imager Network ({FIN}): A deep neural network for hologram reconstruction with superior external generalization},
  journal = {Light: Science \& Applications},
  volume  = {11},
  pages   = {254},
  year    = {2022}
}

@inproceedings{ronneberger2015,
  author    = {Ronneberger, O. and Fischer, P. and Brox, T.},
  title     = {{U-Net}: Convolutional networks for biomedical image segmentation},
  booktitle = {Medical Image Computing and Computer-Assisted Intervention},
  pages     = {234--241},
  year      = {2015}
}

@inproceedings{he2016,
  author    = {He, K. and Zhang, X. and Ren, S. and Sun, J.},
  title     = {Deep residual learning for image recognition},
  booktitle = {Proceedings of the IEEE Conference on Computer Vision and Pattern Recognition},
  pages     = {770--778},
  year      = {2016}
}

@article{chen2018,
  author  = {Chen, L.-C. and Papandreou, G. and Kokkinos, I. and Murphy, K. and Yuille, A. L.},
  title   = {{DeepLab}: Semantic image segmentation with deep convolutional nets, atrous convolution, and fully connected {CRFs}},
  journal = {IEEE Transactions on Pattern Analysis and Machine Intelligence},
  volume  = {40},
  number  = {4},
  pages   = {834--848},
  year    = {2018}
}

@inproceedings{woo2018,
  author    = {Woo, S. and Park, J. and Lee, J.-Y. and Kweon, I. S.},
  title     = {{CBAM}: Convolutional block attention module},
  booktitle = {Proceedings of the European Conference on Computer Vision},
  pages     = {3--19},
  year      = {2018}
}

@book{mardia2000,
  author    = {Mardia, K. V. and Jupp, P. E.},
  title     = {Directional Statistics},
  publisher = {Wiley},
  address   = {Chichester, U.K.},
  year      = {2000}
}

@article{cremers2018,
  author  = {Cremers, J. and Klugkist, I.},
  title   = {One direction? A tutorial for circular data analysis using {R} with examples in cognitive psychology},
  journal = {Frontiers in Psychology},
  volume  = {9},
  pages   = {2040},
  year    = {2018}
}

@misc{liu2026,
  author       = {Liu, C. Y. and Cheng, J. and Chen, C.-C. and Shu, S. F.},
  title        = {Circular phase representation and geometry-aware optimization for ptychographic image reconstruction},
  howpublished = {arXiv:2604.26664},
  year         = {2026}
}

@article{guizar2008,
  author  = {Guizar-Sicairos, M. and Thurman, S. T. and Fienup, J. R.},
  title   = {Efficient subpixel image registration algorithms},
  journal = {Optics Letters},
  volume  = {33},
  number  = {2},
  pages   = {156--158},
  year    = {2008}
}

@article{yang2017,
  author  = {Yang, H. and others},
  title   = {Electron ptychographic phase imaging of light elements in crystalline materials using Wigner distribution deconvolution},
  journal = {Ultramicroscopy},
  volume  = {180},
  pages   = {173--179},
  year    = {2017}
}

@article{bangun2023,
  author  = {Bangun, A. and Baumeister, P. F. and Clausen, A. and Weber, D. and Dunin-Borkowski, R. E.},
  title   = {Wigner distribution deconvolution adaptation for live ptychography reconstruction},
  journal = {Microscopy and Microanalysis},
  volume  = {29},
  number  = {3},
  pages   = {994--1008},
  year    = {2023}
}

@article{ongie2020,
  author  = {Ongie, G. and Jalal, A. and Metzler, C. A. and Baraniuk, R. G. and Dimakis, A. G. and Willett, R. and Soltanolkotabi, M.},
  title   = {Deep learning techniques for inverse problems in imaging},
  journal = {IEEE Journal on Selected Areas in Information Theory},
  volume  = {1},
  number  = {1},
  pages   = {39--56},
  year    = {2020}
}

@article{babu2023,
  author  = {Babu, A. V. and others},
  title   = {Deep learning at the edge enables real-time streaming ptychographic imaging},
  journal = {Nature Communications},
  volume  = {14},
  pages   = {7059},
  year    = {2023}
}

@article{pan2023,
  author  = {Pan, X. and others},
  title   = {An efficient ptychography reconstruction strategy through fine-tuning of large pre-trained deep learning model},
  journal = {iScience},
  volume  = {26},
  number  = {12},
  pages   = {108420},
  year    = {2023}
}

@article{chang2023,
  author  = {Chang, D. J. and others},
  title   = {Deep-learning electron diffractive imaging},
  journal = {Physical Review Letters},
  volume  = {130},
  number  = {1},
  pages   = {016101},
  year    = {2023}
}

@article{friedrich2023,
  author  = {Friedrich, T. and Yu, C.-P. and Verbeeck, J. and Van Aert, S.},
  title   = {Phase object reconstruction for {4D-STEM} using deep learning},
  journal = {Microscopy and Microanalysis},
  volume  = {29},
  number  = {1},
  pages   = {395--409},
  year    = {2023}
}

@article{molina2015vibrational,
  author  = {Molina-S{\'a}nchez, Alejandro and Hummer, Kerstin and Wirtz, Ludger},
  title   = {Vibrational and Optical Properties of {MoS$_2$}: From Monolayer to Bulk},
  journal = {Surface Science Reports},
  volume  = {70},
  number  = {4},
  pages   = {554--586},
  year    = {2015},
  doi     = {10.1016/j.surfrep.2015.10.001}
}

@article{maeland1964lattice,
  author  = {Maeland, Arnulf and Flanagan, Ted B.},
  title   = {Lattice Spacings of Gold--Palladium Alloys},
  journal = {Canadian Journal of Physics},
  volume  = {42},
  number  = {11},
  pages   = {2364--2366},
  year    = {1964},
  doi     = {10.1139/p64-213}
}

@misc{yue2026realtime,
  author       = {Yue, H. and Chen, C.-C. and Hsiao, C.-N. and Cheng, J. and Liu, Y. and Liao, X. Z. and Shu, Steve F.},
  title        = {Real-Time Atomic-Resolution Electron Phase Imaging without Probe Calibration via Ptychography-Supervised Learning},
  howpublished = {arXiv:2609.25684},
  year         = {2026}
}
}

\appendix


\section{Zero-Shot Validation on Unseen HEA-NPs}
\label{app:heanp}

\subsection{HEA-NP Dataset and Evaluation Protocol}
\label{app:heanp_setting}

To further examine whether the learned diffraction-to-phase mapping generalizes beyond the two material domains considered in the main experiments, we additionally evaluate the proposed method on a 4D-STEM dataset of high-entropy alloy nanoparticles (HEA-NPs). 
High-entropy alloys are compositionally complex metallic systems containing multiple principal elements.
Compared with conventional single- or few-component materials, such multicomponent compositions can exhibit substantial local chemical and structural variations. The HEA-NP dataset is treated exclusively as an unseen target domain and is not used during model training, model selection, or fine-tuning.

The HEA-NP dataset is included solely as an additional unseen-domain validation. Accordingly, the default 16×16 phase-target size determined from the main experiments is retained without target-specific adjustment.


The HEA-NP dataset contains a $400\times400$ raster scan, corresponding to 160,000 diffraction patterns. Each scan position records a $63\times63$ diffraction-intensity pattern, and the corresponding ePIE reference object has a spatial size of $1023\times1023$ pixels. 

We consider two zero-shot transfer settings: MoS$_2$ $\rightarrow$ HEA-NP and AuPd $\rightarrow$ HEA-NP. For each setting, the model previously trained on the corresponding source material is directly applied to all scan positions of the HEA-NP dataset without retraining or target-domain fine-tuning. The source-domain preprocessing and normalization procedure is retained during inference, and the predicted local phase patches are assembled using the same scan-position-aware Gaussian-window stitching procedure described in Section~3.6.

The quantitative evaluation follows the protocol described in Section~4.1. The ePIE reconstruction of the HEA-NP specimen is used as the reference. Small residual translational offsets between the predicted and reference phase maps are corrected before evaluation. MSE, PSNR, and MS-SSIM are reported over the valid full-field reconstruction region. Two representative regions of interest are additionally shown to examine local structural preservation.

\subsection{Zero-Shot Reconstruction Results}
\label{app:heanp_results}

\begin{figure*}[!t]
    \centering
    \includegraphics[width=0.90\textwidth]{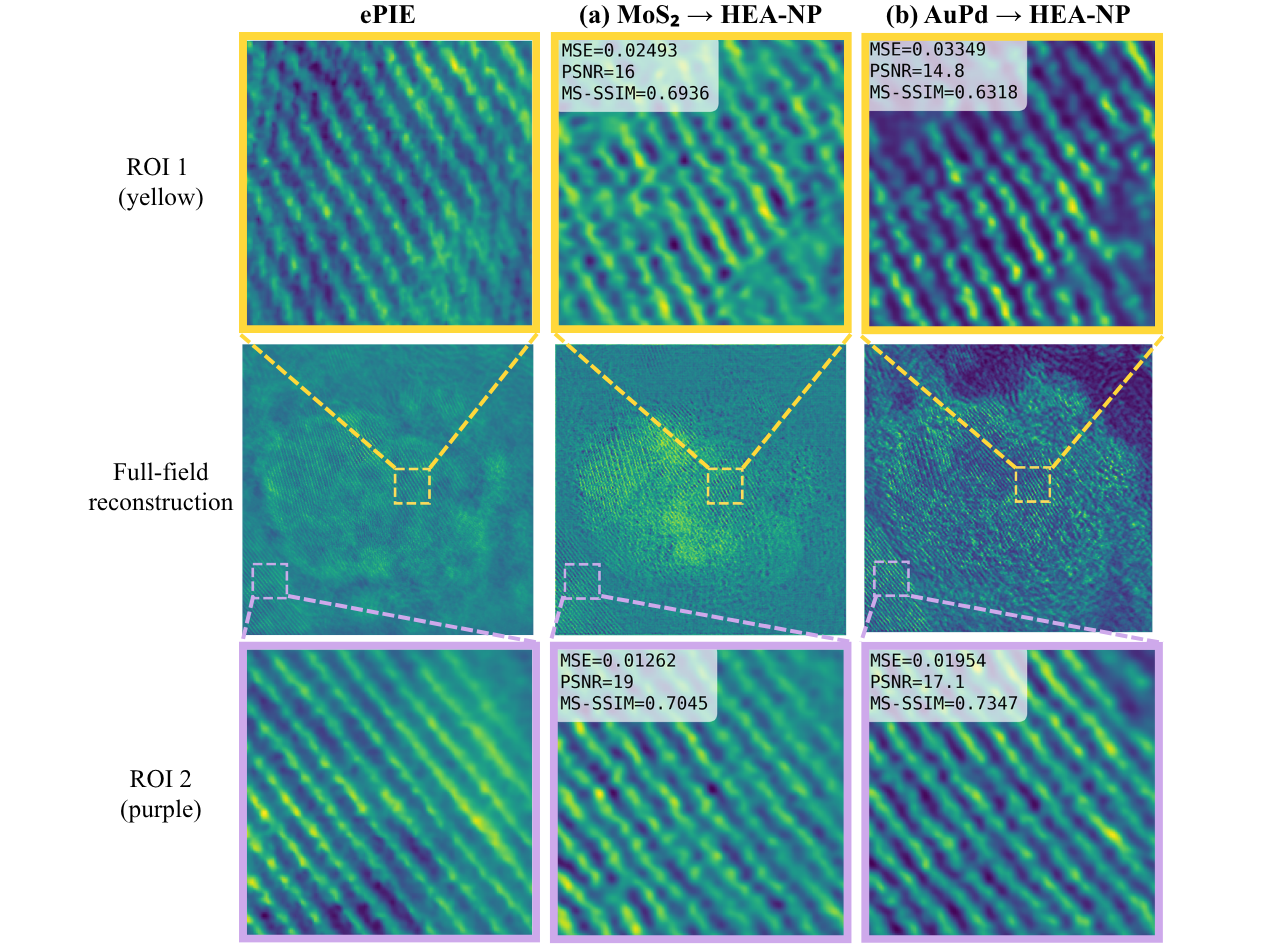}
    \caption{Additional zero-shot reconstruction on the unseen HEA-NP dataset. The ePIE column shows the reference reconstruction. Columns (a) and (b) show predictions obtained using models trained on MoS$_2$ and AuPd, respectively.}
    \label{fig:heanp_zero_shot}
\end{figure*}

Figure~\ref{fig:heanp_zero_shot} presents the zero-shot reconstruction results on the unseen HEA-NP specimen using models trained independently on MoS$_2$ and AuPd. Despite the absence of HEA-NP data during training, both source-domain models recover recognizable local directional phase features observed in the ePIE reference. The representative ROIs further show that both models preserve the dominant local morphology, although source-dependent differences remain in local contrast and structural detail.

Across the representative ROIs, the MoS$_2$-trained model generally shows closer local agreement with the ePIE reference, while the AuPd-trained model remains competitive in structural similarity in selected regions. These regional variations suggest that the source material can influence local reconstruction characteristics, while both models retain the dominant structural features of the unseen HEA-NP specimen.

The full-field results in Table~\ref{tab:heanp_zero_shot} show complementary trends between the two source-domain models rather than a consistent advantage for one model across all metrics. Overall, both independently trained models retain substantial reconstruction capability when directly transferred to a third, previously unseen material domain. These results provide additional evidence that the learned local diffraction-to-phase mapping is not restricted to the two material domains evaluated in the main cross-material experiments.

\begin{table}[!t]
    \centering
    \caption{Full-field zero-shot reconstruction results on the unseen HEA-NP dataset.}
    \label{tab:heanp_zero_shot}
    \resizebox{\columnwidth}{!}{%
    \begin{tabular}{lccc}
        \toprule
        Training $\rightarrow$ Testing
        & MSE $\downarrow$
        & PSNR (dB) $\uparrow$
        & MS-SSIM $\uparrow$ \\
        \midrule
        MoS$_2$ $\rightarrow$ HEA-NP
        & 0.003146
        & 25.23
        & 0.7431 \\
        AuPd $\rightarrow$ HEA-NP
        & 0.004274
        & 25.58
        & 0.7288 \\
        \bottomrule
    \end{tabular}%
    }
\end{table}


\section{Additional Comparison with MRDM}
\label{app:mrdm}

\subsection{Detailed Setting}
\label{app:mrdm_setting}

We additionally reproduce and compare with the Multi-domain Representation and Domain-Attention Model (MRDM) \cite{yue2026realtime}, a learning-based ptychographic reconstruction model designed to exploit complementary diffraction information through a dual-encoder architecture. Its low-frequency branch uses convolutional feature extraction with multi-scale context aggregation to capture the diffraction envelope and long-range variations, while its high-frequency branch employs a Transformer-based representation with polar positional encoding to model finer interference structures. The two representations are subsequently integrated through domain-attention fusion, allowing the network to adaptively balance information from the two diffraction domains.

\begin{figure*}[!t]
    \centering
    \includegraphics[width=0.90\textwidth]{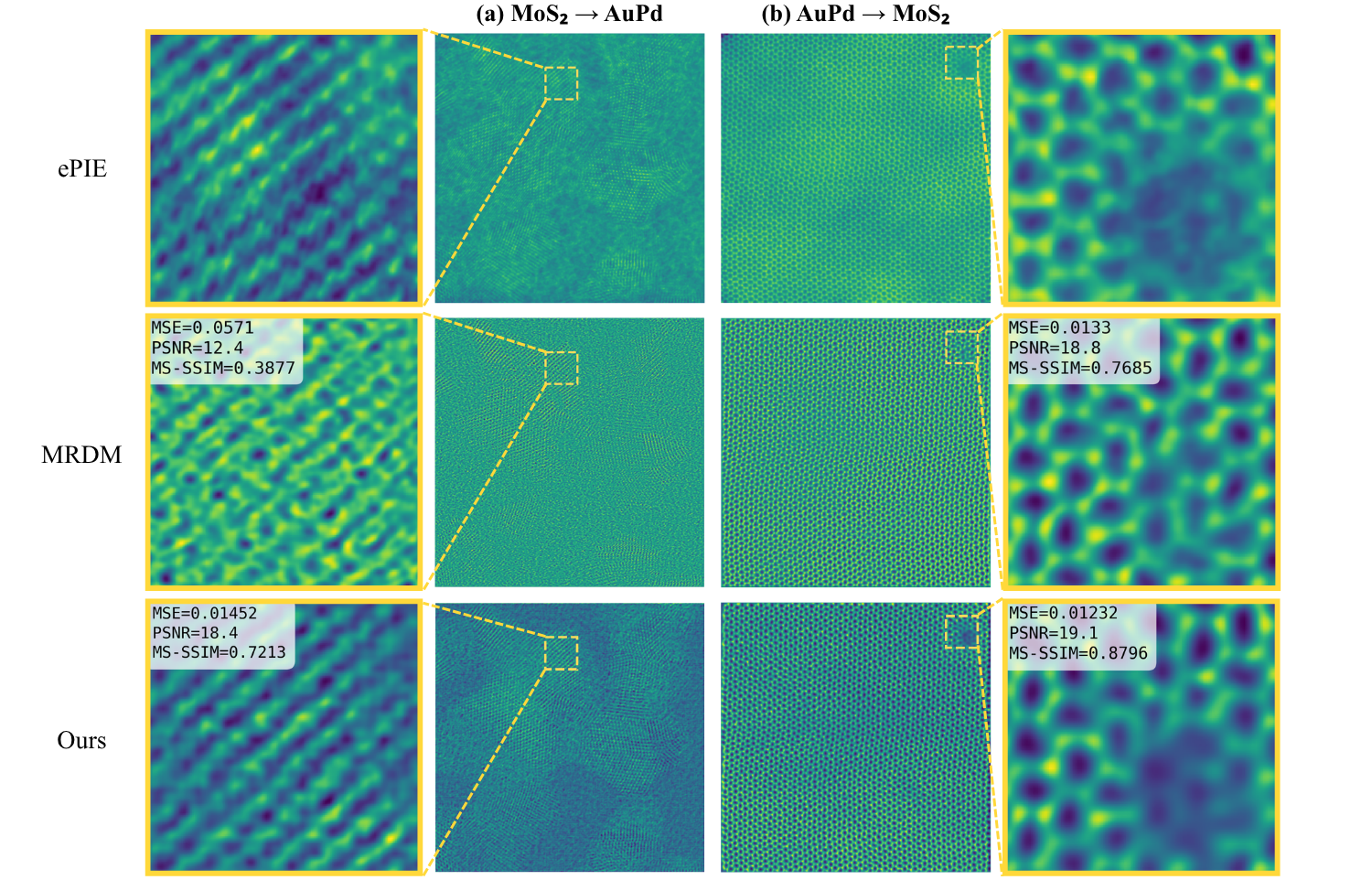}
    \caption{Additional cross-material comparison with MRDM. (a) MoS$_2$ $\rightarrow$ AuPd and (b) AuPd $\rightarrow$ MoS$_2$. Rows show the ePIE reference, MRDM reconstruction, and proposed reconstruction. For each transfer setting, the full-field reconstruction and a representative enlarged region are shown, with the dashed box indicating the location of the enlarged region. The overlaid metrics are computed over the corresponding ROI, while full-field quantitative results are reported in Table~\ref{tab:mrdm_results}.}
    \label{fig:mrdm_cross}
\end{figure*}

Both MRDM and our method perform local prediction from diffraction measurements, but they differ substantially in how diffraction-domain features are represented and fused. We therefore include MRDM as an additional comparison under the same evaluation protocol used throughout this work. The four evaluated settings include AuPd $\rightarrow$ MoS$_2$ and MoS$_2$ $\rightarrow$ AuPd for cross-material transfer, and AuPd $\rightarrow$ AuPd and MoS$_2$ $\rightarrow$ MoS$_2$ for intra-material transfer between independently acquired datasets. The corresponding ePIE reconstructions are used as references. Figures~\ref{fig:mrdm_cross} and~\ref{fig:mrdm_intra} show the full-field reconstructions together with representative enlarged regions, while Table~\ref{tab:mrdm_results} reports the full-field MSE, PSNR, and MS-SSIM. The metrics overlaid on the enlarged regions are computed over the corresponding ROIs.

\subsection{Reconstruction Comparison}
\label{app:mrdm_results}

Figure~\ref{fig:mrdm_cross} presents the cross-material comparisons. For MoS$_2$ $\rightarrow$ AuPd, the proposed method more closely preserves the directional local phase structures observed in the ePIE reference, whereas MRDM exhibits stronger local deviations in the displayed region. For AuPd $\rightarrow$ MoS$_2$, both methods recover the characteristic periodic morphology of the MoS$_2$ specimen, while the proposed reconstruction more consistently preserves the local lattice structure of the reference.

The full-field results in Table~\ref{tab:mrdm_results} support the qualitative observations. The proposed method outperforms MRDM across MSE, PSNR, and MS-SSIM in both cross-material transfer directions, indicating more consistent reconstruction quality under material-domain shift.

\begin{table*}[!t]
    \centering
    \caption{Full-field reconstruction results for the additional comparison with MRDM. Boldface indicates the better result in each setting.}
    \label{tab:mrdm_results}
    \resizebox{0.88\textwidth}{!}{%
    \begin{tabular}{lllccc}
        \toprule
        Evaluation setting
        & Training $\rightarrow$ Testing
        & Method
        & MSE $\downarrow$
        & PSNR (dB) $\uparrow$
        & MS-SSIM $\uparrow$ \\
        \midrule

        \multirow{4}{*}{Cross-material}
        & \multirow{2}{*}{AuPd $\rightarrow$ MoS$_2$}
        & Ours
        & \textbf{0.007819}
        & \textbf{21.24}
        & \textbf{0.6757} \\
        &
        & MRDM
        & 0.036191
        & 14.50
        & 0.5886 \\

        \cmidrule(lr){2-6}

        &
        \multirow{2}{*}{MoS$_2$ $\rightarrow$ AuPd}
        & Ours
        & \textbf{0.005863}
        & \textbf{24.52}
        & \textbf{0.6573} \\
        &
        & MRDM
        & 0.028229
        & 16.46
        & 0.5653 \\

        \midrule

        \multirow{4}{*}{Intra-material}
        & \multirow{2}{*}{AuPd $\rightarrow$ AuPd}
        & Ours
        & \textbf{0.002990}
        & \textbf{29.69}
        & \textbf{0.7295} \\
        &
        & MRDM
        & 0.011664
        & 22.09
        & 0.7048 \\

        \cmidrule(lr){2-6}

        &
        \multirow{2}{*}{MoS$_2$ $\rightarrow$ MoS$_2$}
        & Ours
        & \textbf{0.005261}
        & \textbf{22.95}
        & \textbf{0.8311} \\
        &
        & MRDM
        & 0.058573
        & 12.33
        & 0.5404 \\

        \bottomrule
    \end{tabular}%
    }
\end{table*}

\begin{figure*}[!t]
    \centering
    \includegraphics[width=0.90\textwidth]{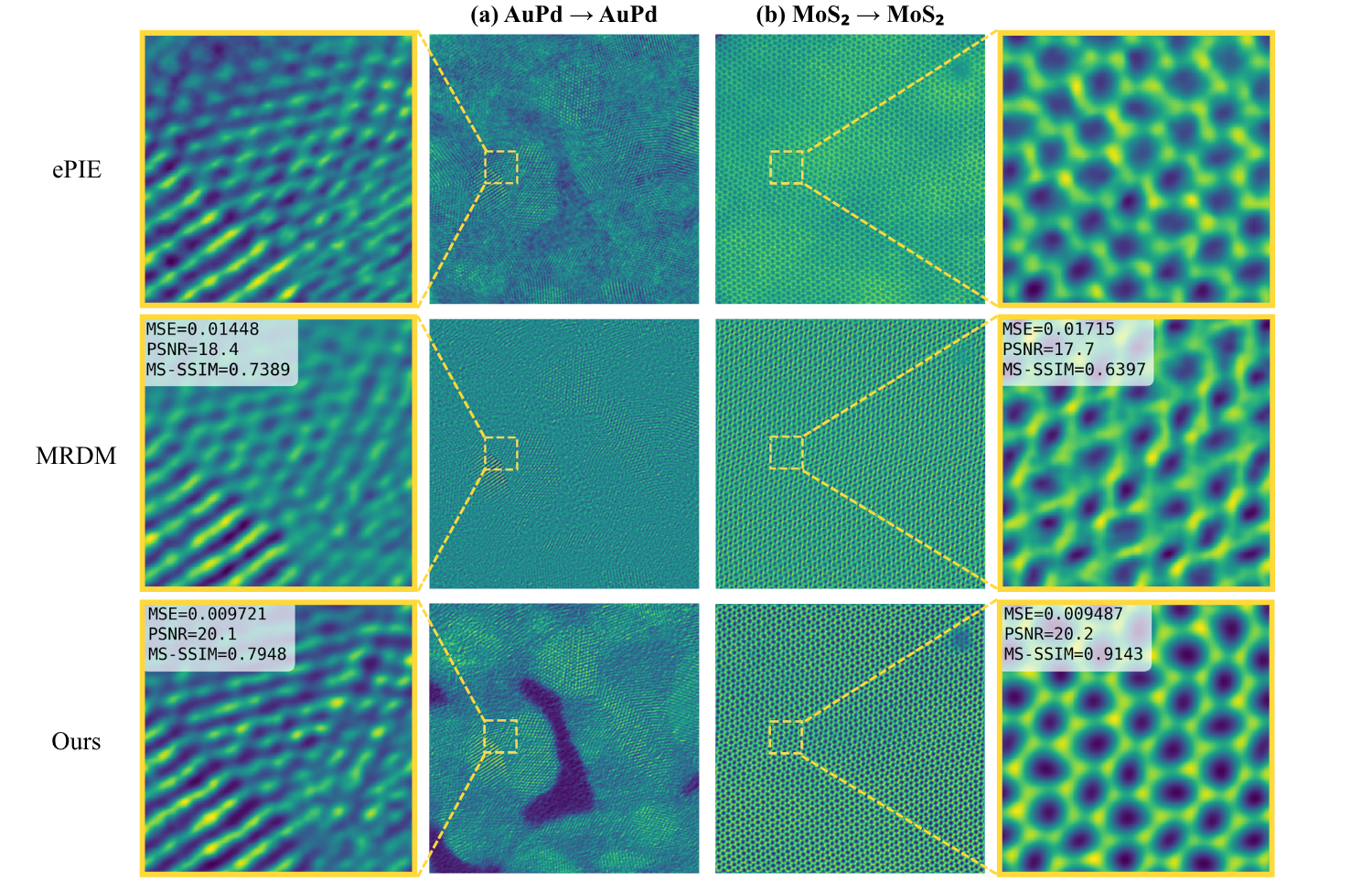}
    \caption{Additional intra-material comparison with MRDM. (a) AuPd $\rightarrow$ AuPd and (b) MoS$_2$ $\rightarrow$ MoS$_2$. Rows show the ePIE reference, MRDM reconstruction, and proposed reconstruction. For each transfer setting, the full-field reconstruction and a representative enlarged region are shown, with the dashed box indicating the location of the enlarged region. The overlaid metrics are computed over the corresponding ROI, while full-field quantitative results are reported in Table~\ref{tab:mrdm_results}.}
    \label{fig:mrdm_intra}
\end{figure*}

Figure~\ref{fig:mrdm_intra} presents the intra-material comparisons between independently acquired datasets of the same material. For AuPd $\rightarrow$ AuPd, both methods recover the dominant crystalline structures, while localized differences remain visible in the reconstructed field. For MoS$_2$ $\rightarrow$ MoS$_2$, both methods preserve the periodic lattice structure, with the proposed reconstruction more closely reproducing the regular local morphology of the ePIE reference.

A similar trend is observed in the intra-material evaluation. The proposed method achieves better full-field results than MRDM across all three metrics for both material domains, suggesting stable reconstruction performance across independently acquired datasets.

Overall, the proposed method achieves more consistent full-field reconstruction quality than MRDM across all four evaluated transfer settings. The advantage is observed in both cross-material transfer and intra-material evaluation, providing additional evidence that the proposed local-to-global framework remains robust under material-domain shift and acquisition-level variation without target-domain fine-tuning.

\end{document}